%% file: main.tex
\PassOptionsToPackage{breaklinks,colorlinks,allcolors=cvprblue}{hyperref}
\documentclass[runningheads]{llncs}

\usepackage{eccv}

\usepackage{orcidlink}

\usepackage{eccvabbrv}

\usepackage{graphicx}
\usepackage{booktabs}

\usepackage[accsupp]{axessibility}  

\definecolor{cvprblue}{rgb}{0.21,0.49,0.74}
\usepackage[breaklinks,colorlinks,allcolors=cvprblue]{hyperref}
\usepackage{algorithm}
\usepackage{algorithmic}
\usepackage{microtype}
\usepackage[most,skins]{tcolorbox} 
\usepackage{caption}
\usepackage{comment}
\usepackage{multirow}
\usepackage{amsmath}
\usepackage{amsfonts}
\usepackage{adjustbox}
\usepackage{makecell}
\usepackage[table]{xcolor}
\usepackage{colortbl}
\usepackage{graphicx}
\usepackage{pifont}
\definecolor{PastaYellow}{RGB}{255,230,190}
\definecolor{GroupColor1}{RGB}{220,230,255}
\definecolor{GroupColor2}{RGB}{220,255,220}
\definecolor{GroupColor3}{RGB}{255,220,200}
\definecolor{GroupColor4}{RGB}{230,220,255}
\definecolor{GroupColor5}{RGB}{210,200,205}
\definecolor{darkgreen}{HTML}{006400}
\definecolor{darkred}{RGB}{139,0,0}
\newcommand{\cmark}{\ding{51}}
\newcommand{\cmarkg}{\textcolor{darkgreen}{\cmark}}
\newcommand{\xmarkr}{\textcolor{red}{\xmark}}
\newcommand{\xmark}{\ding{55}}

\definecolor{AcademicBlue}{HTML}{0055A4} 
\definecolor{ContrastingOrange}{HTML}{E69F00} 
\definecolor{DarkerBlue}{HTML}{002D5A}
 \definecolor{DarkGray}{HTML}{2F2F2F}

\usepackage{newfloat}
\usepackage{listings}
\usepackage{booktabs}
\usepackage{tcolorbox}
\definecolor{CVPRBoxBackground}{rgb}{0.96, 0.94, 0.90} 
\definecolor{CVPRBoxBorder}{rgb}{0.82, 0.75, 0.68}     
\definecolor{CVPRTitle}{rgb}{0.10, 0.10, 0.10} 

\usepackage{tikz}
\usepackage{xspace}
\usepackage{enumitem}

\newcommand{\blocktag}[3]{
  \tikz[baseline=(X.base)]\node[inner sep=2pt, line width=0.5pt,
    draw=#2, fill=#3, rounded corners=2pt, font=\ttfamily\footnotesize]
    (X) {\vphantom{lg}#1};\xspace
}

\definecolor{egoplancolor}{HTML}{1565C0}
\newcommand{\egoplan}{\blocktag{<ego\_plan>}{egoplancolor}{egoplancolor!10}}

\newcommand{\exoverify}{\blocktag{<exo\_verify>}{orange!80!black}{orange!10}}
\newcommand{\answer}{\blocktag{<answer>}{green!50!black}{green!8}}

\makeatletter
\renewcommand\paragraph{\@startsection{paragraph}{4}{\z@}
  {0.8ex \@plus0.1pt \@minus.5ex}
  {-0.2em}
  {\normalfont\normalsize\bfseries}}
\makeatother
\makeatletter

\floatstyle{ruled}
\newfloat{listing}{tb}{lst}{}
\floatname{listing}{Listing}

\lstdefinestyle{cleanpythonstyle}{
    language=Python,
    backgroundcolor=\color{gray!5},
    commentstyle=\color{green!60!black}\itshape,
    keywordstyle=\color{blue!80!black}\bfseries,
    numberstyle=\tiny\color{gray!70},
    stringstyle=\color{red!70!black},
    identifierstyle=\color{black},
    basicstyle=\ttfamily\small,
    breakatwhitespace=false,         
    breaklines=true,                 
    postbreak=\mbox{\textcolor{red}{$\hookrightarrow$}\space},
    captionpos=b,                    
    keepspaces=true,                 
    numbers=left,                    
    numbersep=8pt,
    stepnumber=1,
    firstnumber=1,
    showspaces=false,                
    showstringspaces=false,
    showtabs=false,                  
    tabsize=4,
    frame=tb,
    framerule=0.5pt,
    framesep=5pt,
    xleftmargin=15pt,
    xrightmargin=5pt,
    aboveskip=10pt,
    belowskip=10pt,
    escapeinside={(*@}{@*)},
    mathescape=true,
}

\newcommand{\modelname}{EgoVITA}

\title{EgoVITA: Learning to Plan and Verify for Egocentric Video Reasoning}

\author{Yogesh Kulkarni\orcidlink{0000-0001-9246-3448} \and
Pooyan Fazli\orcidlink{0000-0002-2625-8216}}

\authorrunning{Y.~Kulkarni and P.~Fazli}

\institute{Arizona State University\\
\email{ykulka10@asu.edu, pooyan@asu.edu} \\[2mm]
  \href{https://people-robots.github.io/EgoVITA/}{\textcolor{black}{\texttt{https://people-robots.github.io/EgoVITA/}}}}

\begin{document}

\maketitle

\input{sec/0_abstract}

\input{sec/1_intro}

\input{sec/2_related}

\input{sec/3_method}

\input{sec/4_experiments}

\section*{Acknowledgments}
This research was supported by the National Eye Institute (NEI) of the National Institutes of Health (NIH) under award number R01EY034562.\ The content is solely the responsibility of the authors and does not necessarily represent the official views of the NIH. We acknowledge Research Computing at Arizona State University for providing computing resources.
\bibliographystyle{splncs04}
\bibliography{main}

\input{sec/appendix}

\end{document}

%% file: sec/0_abstract.tex
\begin{abstract}
Egocentric video understanding requires procedural reasoning under partial observability and continuously shifting viewpoints. Current multimodal large language models (MLLMs) struggle with this setting, often generating plausible but visually inconsistent or weakly grounded responses. We introduce \textbf{\modelname{}}, a framework that decomposes egocentric video reasoning into a structured \textit{plan-then-verify} process. The model first generates an \textbf{egocentric plan}: a causal sequence of anticipated actions from a first-person perspective. This plan is then evaluated by an \textbf{exocentric verification} stage that uses third-person reasoning over the same video to verify its spatiotemporal and logical consistency, without exocentric video input.
This decomposition enables cross-perspective feedback without requiring paired ego-exo supervision. To train this reasoning process, we adopt Group Relative Policy Optimization (GRPO) with two dense reward signals: one that grounds anticipated actions in subsequent visual observations and another that reinforces consistent third-person verification. \modelname{} achieves state-of-the-art performance on egocentric reasoning benchmarks, outperforming Qwen2.5-VL-7B by $\mathbf{+7.7}$ on EgoBlind and $\mathbf{+4.4}$ on EgoOrient, while maintaining strong generalization on exocentric video tasks with only $52k$ training samples. 
\keywords{Egocentric Video Understanding \and Procedural Reasoning \and Reinforcement Learning}
\end{abstract}

%% file: sec/1_intro.tex
\section{Introduction}

Egocentric, or first-person, videos play a central role in applications such as life-logging, personal memory assistants~\cite{yang2025egolife}, and assistive technologies~\cite{kim2025guidedog, xiao2025egoblind}. Unlike exocentric footage, egocentric videos are captured from the actor’s viewpoint, resulting in continuous camera motion, partial observability, and frequent occlusions. Effective egocentric understanding therefore requires procedural reasoning to predict how actions unfold, how object states evolve, and how goals are achieved over time. Despite recent advances, multimodal large language models (MLLMs) remain weak in this setting. They often produce plausible yet incorrect descriptions~\cite{seth2025egoillusion}, struggle with spatiotemporal reasoning~\cite{plizzari2025omnia, liang2025finegrained}, and fail to maintain coherent object and goal representations~\cite{yuan2025eoc,li2025egotom, peng2025in, sun2025visual} across time.

These shortcomings reveal a key gap: current MLLMs lack explicit mechanisms to reason about how actions and object states evolve over time in first-person video. While supervised fine-tuning (SFT) on egocentric datasets can improve in-domain performance, it often degrades accuracy on standard exocentric benchmarks, highlighting the challenge of generalizing across viewpoints. Recent methods~\cite{egothinker} attempt to overcome this by scaling to massive datasets, but rely on sparse, rule-based rewards that provide limited guidance for procedural reasoning and still suffer from catastrophic forgetting of exocentric performance.

\input{figures/main_figure}

We hypothesize that these failures stem not from insufficient visual information but from the lack of explicit mechanisms to validate first-person predictions against broader scene constraints. When a model plans from an egocentric view (e.g., ``I will reach for the cup''), it must also assess whether that plan is physically plausible (e.g., ``a cup is within arm’s reach on the counter''). Such validation requires reasoning about scene layout, object affordances, and spatial relationships from a complementary third-person perspective. Recent work suggests that MLLMs can reason across viewpoints, demonstrating perspective-taking and perspective transformation abilities~\cite{lee2025perspective, bu2025walk, actial}. For instance, given a first-person video, a model may reinterpret the scene from an external observer’s perspective (e.g., ``A person is reaching toward a cup on the counter''), indicating the ability to shift between egocentric and exocentric descriptions. 

Building on this, we introduce \textbf{\modelname{}} (\textbf{\underline{Ego}}centric \textbf{\underline{V}}ideo \textbf{\underline{I}}ntelligence and \textbf{\underline{T}}hinking \textbf{\underline{A}}gent), a framework that decomposes egocentric reasoning into two distinct stages: (1) an \textbf{egocentric planning} stage that predicts a causal sequence of first-person actions, and (2) an \textbf{exocentric verification} stage that audits this plan using third-person reasoning over the same egocentric video, without requiring exocentric input.
This separation provides two advantages. First, the planning stage can specialize in egocentric procedural reasoning. Second, by retaining a dedicated verification component grounded in third-person reasoning, the framework preserves exocentric performance and mitigates catastrophic forgetting without requiring synchronized ego-exo video pairs~\cite{jung2025egoexo}.

To train this plan-then-verify framework, we adopt reinforcement learning to provide trajectory-level supervision for multi-step reasoning. Specifically, we use Group Relative Policy Optimization (GRPO)~\cite{grpo}, which evaluates multiple candidate trajectories per input and updates the policy based on their relative quality. Within this framework, we introduce two dense reward signals: Anticipatory Cross-Modal Grounding (ACMG), which grounds anticipated actions by encouraging their visual and temporal consistency with subsequent observations, and a \textit{confidence reward}, which reinforces third-person validation of the predicted plan. In summary, our contributions are as follows:

\begin{enumerate}
   \item We propose \modelname{}, a framework that decomposes egocentric reasoning into an egocentric planner and an exocentric verifier, introducing explicit feedback between first-person prediction and third-person validation.
   \item We introduce two dense reward mechanisms, ACMG and a confidence reward, that promote anticipatory grounding of predicted actions and consistent third-person verification.
   \item We demonstrate that \modelname{} achieves strong performance across both egocentric and exocentric benchmarks, improving Qwen2.5-VL-7B by \textbf{+7.7} on EgoBlind and \textbf{+4.4} on EgoOrient, while maintaining competitive performance on exocentric video tasks within a single unified model.
\end{enumerate}

%% file: figures/main_figure.tex
\begin{figure}[!t]
    \centering
    \includegraphics[width=\columnwidth]{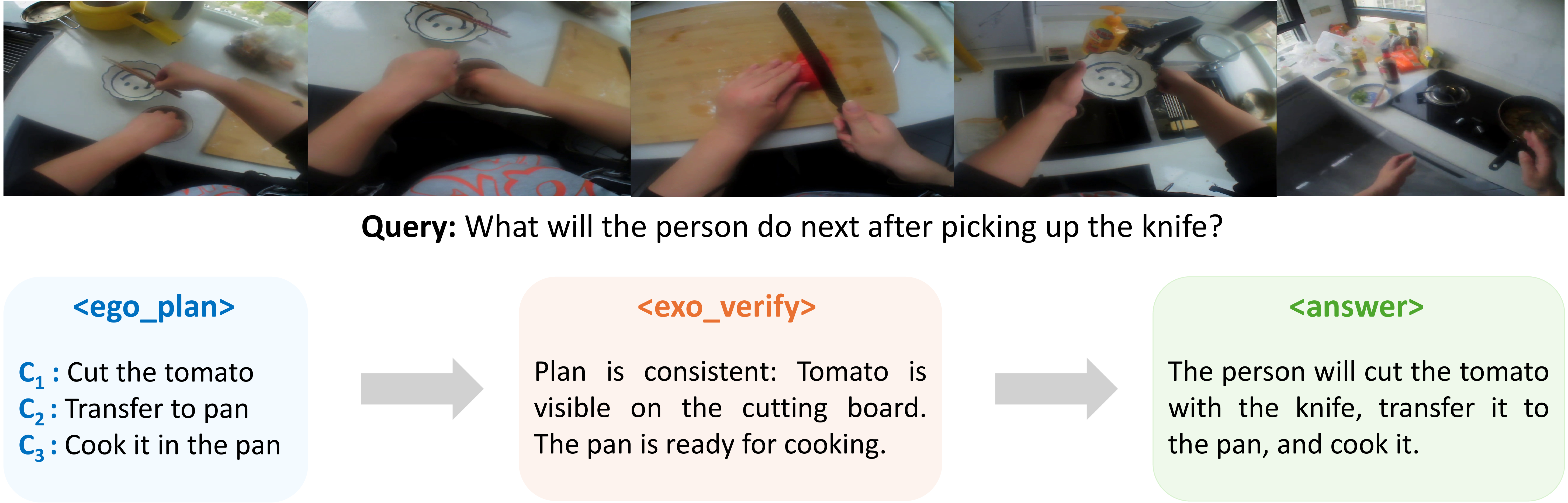}
\caption{\textbf{\modelname{} Framework.} Given a video and query, the model first generates an egocentric plan, then verifies it through third-person reasoning over the same video before producing a final answer.}
\label{fig:training_loop}
    \label{fig:main}
\end{figure}

%% file: sec/2_related.tex
\section{Related Work}

\paragraph{Egocentric Video Understanding.}
Egocentric video reasoning is uniquely challenging due to rapid camera motion, frequent object occlusions~\cite{liang2025finegrained, perrett2025hd}, and the need to infer the intentions of an unobserved, active person~\cite{egothinker, sun2025visual}. 
As a result, recent MLLM research has focused mainly on creating \emph{benchmarks} that highlight specific cognitive failures rather than developing new \emph{methods}. These failures include deficits in long-term memory and procedural reasoning~\cite{ye2024mm, yang2025egolife, tian2025egor1}, fundamental temporal understanding~\cite{plizzari2025omnia, videopasta}, spatial orientation~\cite{jung2025is, gholami2025spatial}, and intention or Theory of Mind grounding~\cite{sun2025visual, peng2025in, li2025egotom}. Models also struggle with fine-grained perception of dynamic object states~\cite{yuan2025eoc} or scene-text~\cite{zhou2025egotextvqa, videosavi}, and often show hallucinations~\cite{seth2025egoillusion} or provide unsafe answers in assistive contexts~\cite{xiao2025egoblind, kim2025guidedog}.
The few methods proposed, such as EgoThinker~\cite{egothinker} and EgoVLM~\cite{vinod2025egovlm}, employ reinforcement learning~\cite{avatar}. However, their reliance on simple, sparse rewards for output format and correctness limits long-term procedural reasoning and often causes catastrophic forgetting of general exocentric knowledge. To address this, we introduce a novel framework that leverages dense, predictive rewards to teach complex egocentric procedures while preserving exocentric generalization, tackling a key challenge in ego-exo knowledge transfer~\cite{zhang2025exo2ego, luo2025viewpoint, huang2025sound}.

\paragraph{Ego-Exo Video Understanding.} 
Applying MLLMs trained on large corpora of exocentric images and videos to egocentric video faces a fundamental \textit{ego-exo gap}~\cite{he2025egoexobench}. This domain shift introduces several challenges, including viewpoint disparity that disrupts feature alignment~\cite{luo2025viewpoint}, difficulty in matching objects across views~\cite{murlabadia2025omama}, and complications in relative spatial reasoning~\cite{gholami2025spatial}. Prior work attempts to bridge this gap by learning view-invariant representations~\cite{luo2025viewpoint}, using audio as a ``sound bridge''~\cite{huang2025sound}, or employing SFT pipelines for direct knowledge transfer~\cite{zhang2025exo2ego, xu2025egoexo}. While these approaches improve cross-view alignment, they struggle to teach dynamic procedural reasoning and often suffer from catastrophic forgetting of exocentric knowledge. \modelname{} overcomes these limitations by explicitly separating reasoning into an egocentric planning phase and an exocentric verification phase. This design enables the model to learn robust first-person understanding while using verification to preserve third-person generalization. Unlike prior work~\cite{jung2025egoexo} that relies on synchronized ego-exo video pairs, \modelname{} achieves scalable cross-view consistency without requiring paired data.

%% file: sec/3_method.tex
\section{EgoVITA}

We introduce \modelname{}, a reinforcement learning framework that decomposes egocentric video reasoning into two complementary components: an \textit{egocentric planner} that performs procedural reasoning by predicting first-person action sequences, and an \textit{exocentric verifier} that evaluates these plans from a third-person perspective for visual and logical consistency. 
We train \modelname{} in two phases. We first apply supervised fine-tuning (SFT) to initialize the structure of plan-then-verify. We then optimize the model using Group Relative Policy Optimization (GRPO) with two dense reward functions: Anticipatory Cross-Modal Grounding (ACMG), which encourages visual and temporal alignment between anticipated actions and subsequent visual observations, and a confidence reward, which reinforces consistent third-person validation. An overview of the framework is shown in Figure~\ref{fig:main}.

\subsection{Problem Formulation}

We formulate egocentric video reasoning as a sequential decision-making problem, 
where a multimodal large language model (MLLM) acts as a policy $\pi_\theta$ 
parameterized by weights $\theta$. 
Given a video $V$ and a text prompt $X$, the policy autoregressively generates a 
\emph{reasoning trajectory}
\begin{equation}
Y = \langle y_1, \dots, y_T \rangle, 
\quad y_t \sim \pi_\theta(y_t \mid V, X, y_{<t}),
\label{eq:autoregressive}
\end{equation}
where each $y_t$ is a generated token. We structure the reasoning trajectory $Y$ into three contiguous segments:
\[
Y = \big( Y^{\text{plan}},\, Y^{\text{verify}},\, Y^{\text{ans}} \big),
\]
corresponding to the egocentric plan \egoplan, 
the exocentric verification \exoverify, 
and the final answer block \answer, respectively.

The quality of a reasoning trajectory $Y$ is evaluated by a composite reward 
function $R(Y)$, which integrates signals encouraging alignment with future 
visual observations and consistency of third-person validation. The objective is to learn parameters $\theta$ that maximize the expected reward:
\begin{equation}
    \theta^* = \arg\max_\theta 
    \; \mathbb{E}_{Y \sim \pi_\theta(\cdot \mid V,X)} [R(Y)].
\label{eq:objective}
\end{equation}

\subsection{Reasoning Decomposition}
\label{sec:decomposition}

\paragraph{Egocentric Planning.}
Given a video $V$ and query $X$, the model first generates an \egoplan: a temporally ordered sequence of action clauses $\{c_1, c_2, \dots, c_K\}$ describing the anticipated actions of the camera wearer (e.g., 1.~Cut the tomato. 2.~Transfer to pan.). Each clause describes an anticipated action whose outcome can be validated against subsequent visual observations. By generating such procedural plans, the model is encouraged to reason about action ordering, object state transitions, and temporal dependencies from a first-person perspective.

\paragraph{Exocentric Verification.}
The generated plan is then audited by an \exoverify stage, which reasons about the scene from a third-person perspective. Operating on the same video, it evaluates the egocentric plan without requiring exocentric video or paired ego-exo data. Given the egocentric plan, the model assesses whether the predicted actions are consistent with scene layout, object affordances, and observable context (e.g., ``The knife is visible on the counter. The cutting board is within reach. Plan is consistent.''). This verification step evaluates the plausibility of the planned actions under a complementary perspective, rather than relying solely on first-person prediction.

\paragraph{Answer.}
Finally, the model produces a final \answer conditioned on both the egocentric plan and its exocentric verification. This decomposition grounds the response in both first-person procedural reasoning and third-person validation.

\subsection{Training}
\label{sec:training}

\modelname{} is trained in two phases. First, SFT initializes the model to produce plan-verify reasoning trajectories. Second, GRPO further optimizes the policy using dense reward signals. We describe each phase below.

\paragraph{Stage I: SFT}
\label{sec:sft}
We first apply SFT to initialize the policy to generate egocentric plans \egoplan, exocentric verifications \exoverify, and final answers \answer under cross-entropy supervision.

We rely on human-annotated datasets as supervision. For egocentric planning, we use EgoProceL~\cite{egoprocel}, which provides time-aligned action annotations labeled by human annotators for egocentric videos. We prompt Qwen2.5-VL-72B~\cite{qwen2_5vl} to generate plan sequences conditioned on these annotations (prompt in Supplementary). For exocentric verification, we construct a dataset based on HD-EPIC~\cite{perrett2025hd}, which contains human-labeled (action - timestamp) video annotations. Qwen2.5-VL-72B generates third-person chain-of-thought rationales grounded in these annotations (prompt in Supplementary). All generated reasoning traces are filtered and verified using Seed1.5-VL~\cite{guo2025seed1} to ensure quality.
Both datasets are merged into a unified corpus $D$ and trained with standard cross-entropy:
\begin{equation}
   \mathcal{L}_{\mathrm{SFT}}(\theta) = - \mathbb{E}_{(V, X, Y^*) \sim D} \left[ \log \pi_\theta(Y^* \mid V, X) \right].
\label{eq:sft_loss}
\end{equation}

\noindent where $(V, X, Y^*)$ denotes a video, prompt, and ground-truth reasoning sequence. 

\paragraph{Stage II: GRPO}
\label{sec:grpo}
After SFT, we further optimize the policy $\pi_{\theta}$ using GRPO, which performs trajectory-level policy updates based on relative reward comparisons. In egocentric video reasoning, multiple plausible reasoning trajectories may exist for a given input due to scene ambiguity and viewpoint dynamics. GRPO addresses this by sampling a group of candidate trajectories per input and updating the policy according to their relative rewards. For each video-prompt pair $(V, X)$, the policy $\pi_\theta$ generates a group of $k=8$ rollouts $\{Y_1, \dots, Y_k\}$, where each trajectory includes a full \egoplan, \exoverify, and \answer segments. Each $Y_i$ is scored using a composite reward function $R(Y_i)$, which integrates alignment with future visual observations and consistency of third-person validation.

\vspace{0.1cm}
\noindent\textit{GRPO Objective.}
Group-level rewards $\{R(Y_i)\}_{i=1}^k$ are normalized to compute relative advantages:
\begin{equation}
\hat{A}(Y_i) = \frac{R(Y_i) - \mu_R}{\sigma_R + \epsilon},
\label{eq:advantage}
\end{equation}
where $\mu_R$ and $\sigma_R$ are the mean and standard deviation of the rewards within the sampled group. The GRPO objective is defined as:

\begin{equation}
\begin{split}
\mathcal{J}_{\mathrm{GRPO}}(\theta) = \mathbb{E}_{(V,X)} \Bigg[ \frac{1}{k} \sum_{i=1}^k \bigg( & \min \left( r_i(\theta) \hat{A}(Y_i), \text{clip}\left(r_i(\theta), 1{-}\epsilon, 1{+}\epsilon\right) \hat{A}(Y_i) \right) \\
& - \beta_{\mathrm{KL}} \log \frac{\pi_\theta(Y_i \mid V, X)}{\pi_{\mathrm{ref}}(Y_i \mid V, X)} \bigg) \Bigg],
\end{split}
\label{eq:grpo_objective}
\end{equation}

\noindent where the probability ratio is

\begin{equation}
r_i(\theta) = \frac{\pi_\theta(Y_i \mid V, X)}{\pi_{\theta_{\mathrm{old}}}(Y_i \mid V, X)}.
\end{equation}

\noindent and $\beta_{\mathrm{KL}}$ limits deviation from the reference policy $\pi_{\mathrm{ref}}$ (initialized from SFT).

\paragraph{Composite Reward ($R$).}
The reward is a weighted sum of four components:
\begin{equation}
R = w_f R_{\mathrm{format}} + w_a R_{\mathrm{answer}}
+ w_g R_{\mathrm{ACMG}} + w_c R_{\mathrm{confidence}},
\label{eq:total_reward}
\end{equation}
\noindent where:
\begin{enumerate}
    \item \textbf{Format Reward ($R_{\mathrm{format}}$)} is a sparse binary signal that checks for correct tag usage and output structure,
    \item \textbf{Answer Reward ($R_{\mathrm{answer}}$)} is a sparse reward measuring alignment between the generated answer and the ground-truth label,
    \item \textbf{ACMG Reward ($R_{\mathrm{ACMG}}$)} is a dense reward that promotes alignment between predicted plan clauses and future visual frames, encouraging temporal and visual grounding, and
\item \textbf{Confidence Reward ($R_{\mathrm{confidence}}$)} is a dense reward that reinforces consistent third-person validation of the predicted plan.
\end{enumerate}
\noindent Sparse rewards ($R_{\mathrm{format}}$, $R_{\mathrm{answer}}$) provide binary pass/fail scores based on output structure and final correctness, while dense rewards ($R_{\mathrm{ACMG}}$, $R_{\mathrm{confidence}}$) provide continuous, fine-grained feedback that evaluates the quality of intermediate reasoning within the trajectory.

\input{figures/acmg_and_temporal_s}

\paragraph{Anticipatory Cross-Modal Grounding (ACMG).}
\label{sec:acmg}
To encourage egocentric plans to align with future visual observations, we introduce the ACMG reward. This dense reward measures the visual and temporal alignment between each plan clause and subsequent observations in the video (Figure~\ref{fig:acmg}). For each clause $c_i$, we extract its final text hidden state $h_i^{\mathrm{text}}$ from the language decoder of the MLLM and project it into the visual embedding space via a small trainable MLP, referred to as the Anticipation Head. The output of this projection is a predicted future visual embedding, denoted as $\hat{v}_{i} = \mathrm{MLP}(h_i^{\mathrm{text}})$. 

Let $v_{t+n}$ denote the visual embedding of frame $t+n$, extracted from the visual encoder of the MLLM. We consider the next $N=16$ frames following timestamp $t$ as a temporal horizon. This window captures near-future action outcomes while limiting influence from distant events. The ACMG reward for each clause $c_i$ is defined as the maximum cosine similarity between the predicted embedding and the embeddings of these future frames (Figure~\ref{fig:acmg_temporal}):

\begin{equation}
    R_{\mathrm{ACMG}}({c_i}) \;=\; \max_{{n} \in \{{1},\dots,{N}\}} \frac{{\hat{v}_{i}}\cdot {v_{t+n}}}{\|{\hat{v}_{i}}\|\,\|{v_{t+n}}\|}.
\label{eq:acmg}
\end{equation}
\noindent 

\noindent The $\max$ operator selects the frame that best aligns with the predicted action, making the reward robust to small temporal offsets and variable action durations. The trajectory-level ACMG reward is computed by averaging across clauses:

\begin{equation}
R_{\mathrm{ACMG}} = \frac{1}{K} \sum_{i=1}^{K} R_{\mathrm{ACMG}}(c_i).
\end{equation}

\noindent Ablation studies in the Supplementary Material validate the choice of the $\max$ operator and the temporal window size $N=16$.

\subsubsection{Confidence Reward.}
\label{sec:confidence}

To improve the quality of \exoverify, we introduce a confidence reward 
$R_{\mathrm{confidence}}$ that applies a relative preference signal over candidate 
verification trajectories.

\vspace{0.1cm}
\noindent\textit{Teacher-guided phase.} During the first 200 RL steps, each of the $k=8$ rollout verifications is compared to a teacher-generated reference verification. The confidence reward encourages the policy to assign a higher likelihood to the teacher output relative to its own rollout:

\begin{equation}
R_{\mathrm{confidence}}
=
\gamma
\Big[
\log \pi_\theta\!\left(y^{T} \mid V, X, Y^{\text{plan}}\right)
-
\log \pi_\theta\!\left(y^{P} \mid V, X, Y^{\text{plan}}\right)
\Big],
\label{eq:confidence_teacher}
\end{equation}

\noindent where $\gamma$ is a scaling coefficient controlling the strength of the preference signal, and $y^{T}$ and $y^{P}$ denote the teacher and policy verification outputs, respectively. We use Qwen2.5-VL-72B~\cite{qwen2_5vl} as a teacher model that generates exocentric verification rationales conditioned on the egocentric plan. 
This warm-up phase aligns the exocentric verifier with teacher-generated third-person audits conditioned on the predicted first-person plan.

\vspace{0.1cm}
\noindent\textit{Self-ranking phase.} After the warm-up period, each rollout verification is scored using the composite reward (Eq.~\ref{eq:total_reward}). The highest-scoring verification $y^{\mathrm{chosen}}$ and lowest-scoring verification $y^{\mathrm{rejected}}$ are compared using a log-probability preference objective:

\begin{equation}
    R_{\text{confidence}} = 
    \begin{cases}
        \gamma \cdot (\log \pi_{\theta}(y_{\text{chosen}} \mid V, X, Y^{\text{plan}}) \\
        \quad - \log \pi_{\theta}(y_{\text{rejected}} \mid V, X, Y^{\text{plan}})), & \text{if scores differ;} \\
        1, & \text{otherwise.}
    \end{cases}
\end{equation}

\noindent This update increases the likelihood of third-person audits that better reconcile the predicted first-person plan with first-person visual observations.

\subsection{Exocentric Regularization}
\label{sec:regularization}

Training solely on egocentric data can lead to catastrophic forgetting, reducing performance on exocentric tasks. To mitigate this, we periodically interleave GRPO updates with a lightweight exocentric regularization step. Specifically, after every 200 RL iterations, we optimize the model on an auxiliary exocentric VideoQA dataset (MSR-VTT~\cite{xu2016msr}) using a cross-entropy loss $\mathcal{L}_{\mathrm{exo}}$. This loss is combined with the GRPO objective:
\begin{equation}
\mathcal{L}_{\mathrm{total}} = \mathcal{L}_{\mathrm{GRPO}} + \lambda_{\mathrm{exo}}\mathcal{L}_{\mathrm{exo}},
\label{eq:total_loss}
\end{equation}

\noindent where $\lambda_{\mathrm{exo}}$ controls the strength of the exocentric regularization term. We use MSR-VTT because its diverse third-person video content spans a broad range of everyday scenes, providing complementary supervision to counterbalance egocentric specialization. No additional exocentric signals are introduced during RL training. Ablation studies in the Supplementary Material analyze the effect of $\lambda_{\mathrm{exo}}$ and confirm that this regularization preserves exocentric performance while maintaining gains on egocentric reasoning tasks.

%% file: figures/acmg_and_temporal_s.tex
\begin{figure}[!t]
    \centering
    \begin{minipage}{0.48\columnwidth}
        \centering
        \includegraphics[width=\linewidth]{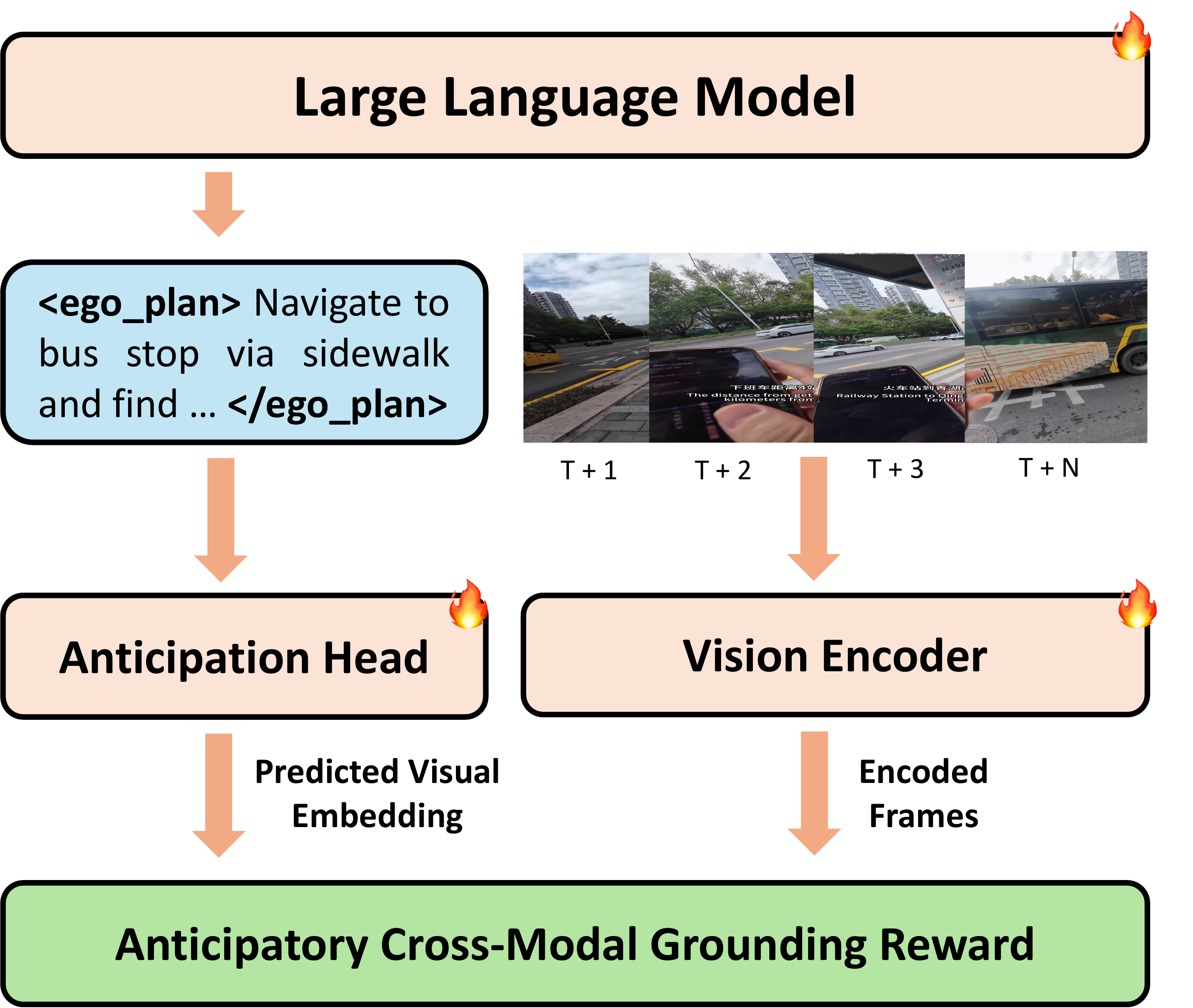}   
        \caption{\textbf{ACMG Mechanism.} Each plan clause is projected into a visual embedding space and matched against the embeddings of the next $N$ frames. The clause-level reward is the maximum cosine similarity over this temporal window.}
        \label{fig:acmg}
    \end{minipage}
    \hfill
    \begin{minipage}{0.48\columnwidth}
        \centering
        \includegraphics[width=\linewidth]{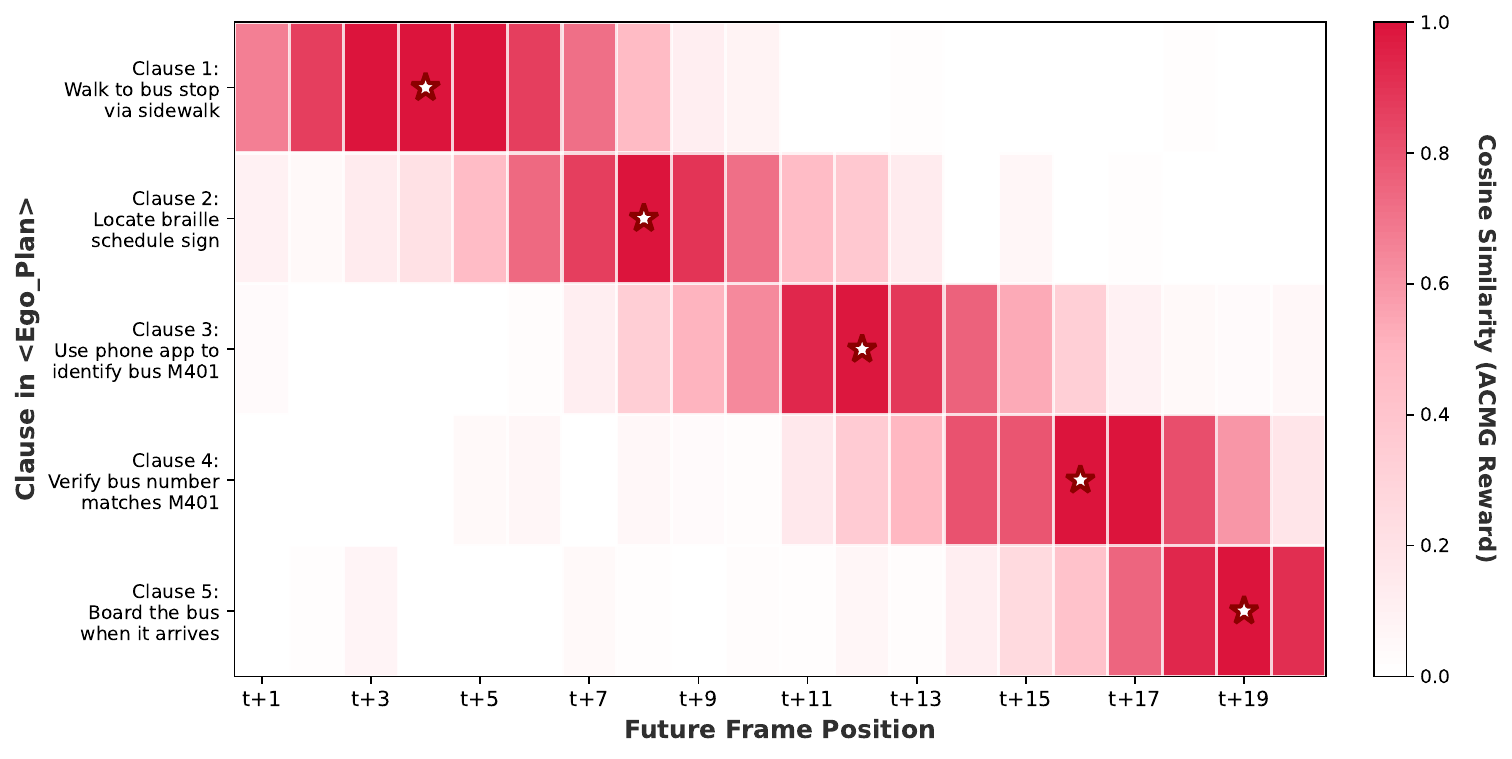}   
        \caption{\textbf{Temporal Alignment.} 
        Heatmap of clause-to-frame similarity, with white stars ($\star$) marking peak matches. The diagonal pattern shows earlier clauses aligning with near-future frames and later clauses with subsequent frames, consistent with anticipatory temporal grounding.}
        \label{fig:acmg_temporal}
    \end{minipage}
\end{figure}

%% file: sec/4_experiments.tex
\section{Experiments and Evaluation}

\paragraph{Implementation Details.} We implement \modelname{} using the open-source MS-SWIFT~\cite{swift} library. In Stage I (SFT), we train on two datasets: EgoProceL~\cite{egoprocel} ($5k$ samples) for egocentric planning and HD-EPIC~\cite{perrett2025hd} ($7k$ samples) for exocentric verification.  Chain-of-thought reasoning traces for both planning and verification are generated once using Qwen2.5-VL-72B~\cite{qwen2_5vl}, conditioned on ground-truth annotations. The SFT model is trained for one epoch with a learning rate of $5e-6$. In Stage II (GRPO), we sample rollouts from  HD-EPIC and EgoIT ($40k$ samples)~\cite{yang2025egolife}. We set the policy temperature $\beta_{\mathrm{KL}} = 0.1$, the confidence scaling coefficient $\gamma=0.1$, and the exocentric regularization coefficient $\lambda_{\text{exo}}=0.05$. We apply \modelname{} to three state-of-the-art MLLMs: Qwen2.5-VL-7B~\cite{qwen2_5vl}, InternVL-3.5-8B~\cite{internvl3_5}, and Qwen3-VL-8B~\cite{qwen3vl}. All models are trained on $8$ NVIDIA L40S GPUs using DoRA~\cite{dora} adapters (\( r = 8\), \(\alpha = 16 \)) on the vision encoder, projector, and language decoder. The Anticipation Head (MLP) is used only during Stage II to compute $R_{\mathrm{ACMG}}$ and is discarded at inference, adding no overhead to the final model. The reward weights are set to $\{w_f, w_a, w_g, w_c\} = \{0.1, 0.3, 0.3, 0.3\}$. The format reward receives a lower weight as it saturates early in training, while the remaining three signals are weighted equally to balance their influence during optimization. At inference, \modelname{} runs at $84.3$ token/s with a time-to-first-token of $0.6$\,s on a single L40S GPU, adding only $0.1$\,s overhead compared to the base Qwen2.5-VL model.

\paragraph{Benchmarks.} We evaluate \modelname{} on several egocentric understanding benchmarks, including EgoBlind~\cite{xiao2025egoblind} (visual assistance for blind users), EgoPlan~\cite{chen2023egoplan} (egocentric task planning), EgoThink~\cite{cheng2024egothink} (first-person perspective thinking), EOC-Bench~\cite{yuan2025eoc} (embodied cognition across past, present, and future), and EgoOrientBench~\cite{jung2025is} (object orientation understanding). We also evaluate on exocentric video understanding benchmarks: MVBench~\cite{mvbench} (temporal perception-to-cognition QA), Video-MME~\cite{videomme} (comprehensive video understanding), LVBench~\cite{wang2025lvbench} (long-video reasoning), and TOMATO~\cite{tomato} (visual temporal reasoning).

\subsection{Results} 
We compare \modelname{} against four baselines:
(1) the original foundation models, 
(2) the same models after SFT,
(3) SFT models further trained with GRPO using only final-answer and format rewards, and
(4) state-of-the-art MLLMs for egocentric video understanding.

\paragraph{\modelname{} improves all foundation models.}
As shown in Table~\ref{tab:ego_benchmarks}, \modelname{} consistently enhances egocentric performance across all three foundation models: Qwen2.5-VL gains $\mathbf{+7.7}$ on EgoBlind, $\mathbf{+3.7}$ on EgoThink, and $\mathbf{+4.4}$ on EgoOrient. InternVL-3.5 improves by $\mathbf{+3.6}$ on EgoBlind and $\mathbf{+3.8}$ on EgoOrient, while the stronger Qwen3-VL achieves gains of $\mathbf{+3.5}$ and $\mathbf{+2.3}$ on the same tasks, suggesting that the improvements generalize across model capacities. These egocentric gains come without hurting
exocentric performance, as all three improve or maintain their scores on exocentric benchmarks (e.g., InternVL gains $\mathbf{+2.7}$ on Video-MME). Furthermore, the self-trained variant using Qwen3-VL-8B (\modelname{}$^{\diamondsuit}$), where both SFT data and teacher confidence signals are generated by the same 8B model, still outperforms the Qwen3-VL baseline on all egocentric benchmarks. This suggests that the improvements are driven by the plan–verify training framework itself, rather than reliance on a larger external teacher.
\input{tables/main_ego}
\input{tables/qwen2vl_based}

\paragraph{Comparison with state-of-the-art.} Work on egocentric video reasoning with MLLMs is limited. EgoThinker~\cite{egothinker} built upon Qwen2-VL~\cite{wang2024qwen2} is the closest related method that enables a controlled backbone comparison. Table~\ref{tab:qwen2vl_based} compares \modelname{} and EgoThinker across standard egocentric reasoning benchmarks and general exocentric tasks, including MMMU~\cite{mmmu}, DocVQA~\cite{docvqa}, MME~\cite{mme}, and VQAv2~\cite{vqa}. Although EgoThinker is trained with approximately \textbf{5M} samples, it shows substantial performance degradation on several exocentric benchmarks (e.g., $\mathbf{-10.5}$ on DocVQA and $\mathbf{-1.8}$ on MMMU), suggesting limited cross-view retention under large-scale egocentric training. In contrast, \modelname{} achieves stronger egocentric performance (e.g., $\mathbf{+5.8}$ over the base model on EgoBlind) using $\mathbf{52k}$ training samples. Furthermore, the proposed dense reward design is associated with improved egocentric reasoning while maintaining competitive performance on exocentric benchmarks (e.g., $\mathbf{+0.6}$ on DocVQA), indicating improved cross-view stability relative to large-scale egocentric fine-tuning.

\input{figures/attention}

\paragraph{RL is essential.}
SFT improves output formatting but provides limited gains in egocentric reasoning and can degrade exocentric performance. For example, Qwen2.5-VL after SFT drops by $\mathbf{-2.0}$ on MVBench and $\mathbf{-2.1}$ on Video-MME, indicating a trade-off between egocentric specialization and exocentric retention. 
In contrast, applying RL through \modelname{} recovers exocentric performance, improving MVBench and Video-MME by $\mathbf{+3.6}$/$\mathbf{+4.5}$ points over SFT while also further increasing EgoBlind by $\mathbf{+2.6}$ points. A similar trend is observed for InternVL-3.5, where RL improves Video-MME by $\mathbf{+6.2}$ points and EgoBlind by $\mathbf{+2.2}$ points over SFT.

\paragraph{Dense rewards matter.}
GRPO using only format and answer rewards provides limited gains over SFT, as these sparse signals primarily optimize final-output correctness. Adding the dense rewards (ACMG and Confidence) yields consistent additional improvements: $\mathbf{+1.8}$ on EgoBlind and $\mathbf{+3.4}$ on EgoOrient for Qwen2.5-VL, and $\mathbf{+1.6}$/$\mathbf{+1.7}$ on the same tasks for InternVL-3.5. These results suggest that grounding plans in future visual observations and reinforcing cross-perspective verification both contribute to improved egocentric reasoning.

\paragraph{Visual Attribution Analysis.}
We apply EAGLE~\cite{eagle} to visualize which regions contribute to action prediction (Figure~\ref{fig:attention}). Qwen2.5-VL shows largely temporally static saliency, with attention frequently concentrated on background elements and limited focus on task-relevant objects. This behavior is accompanied by incorrect action predictions in several cases. In contrast, \modelname{} demonstrates temporally progressive attribution patterns that shift across frames, increasingly focusing on task-critical objects (e.g., garlic, egg bowl) as the action unfolds. This behavior is consistent with the influence of $R_{\mathrm{ACMG}}$, which encourages alignment between predicted actions and future visual observations. The results suggest that the dense grounding signal promotes greater sensitivity to regions that are predictive of subsequent task dynamics, rather than static scene features.

\input{figures/tsne_noun}

\paragraph{Semantic Structure Analysis.} To examine if \modelname{} learns structured visual-linguistic representations, we categorize \egoplan clauses into five procedural steps (Move $\rightarrow$ Locate $\rightarrow$ Action $\rightarrow$ Verify $\rightarrow$ Complete) using GPT-4o~\cite{hurst2024gpt}. Figure~\ref{fig:acmg_tsne} shows a t-SNE projection of the corresponding embeddings, revealing two key properties. First, \textit{between-cluster separation} is visible: embeddings corresponding to different procedural stages form separable clusters, suggesting structured differentiation across task phases. Second, \textit{within-cluster alignment} is strong: the MLP-predicted visual embeddings (red diamonds $\textcolor{red}{\Diamond}$) align closely with embeddings of future frames (green stars $\textcolor{darkgreen}{\star}$), indicating consistent cross-modal alignment between predicted plan clauses and subsequent visual observations.

\paragraph{Robustness to Static Feature Bias.}
To evaluate whether ACMG captures temporal dynamics rather than static background features, we conduct a controlled analysis on 850 EgoPlan~\cite{chen2023egoplan} samples (Figure~\ref{fig:acmg_bias}). For each valid action clause, we pair it with ``Wrong Action'' frames from the same scene (identified by Qwen3-VL-30B-A3B~\cite{qwen3vl}), controlling for static background and object identity while varying only the action. If $R_{\mathrm{ACMG}}$ relied primarily on static scene features, the score distributions would overlap. Instead, correct predictions yield a mean similarity of $\mu=0.68$, whereas same-scene wrong-action pairs drop to $\mu=0.35$ ($\Delta=0.33$). This separation indicates that the Anticipation Head is sensitive to action-dependent visual changes, suggesting that the reward emphasizes temporal variation rather than static scene content.

\input{tables/category_breakdown}

\paragraph{Task-level analysis.}
Table~\ref{tab:category_breakdown} shows that the gains from \modelname{} are concentrated in specific categories rather than arising uniformly from SFT distillation. On EgoBlind, the largest improvements over SFT occur in Safety Warning ($\mathbf{+5.4}$) and Other Resources ($\mathbf{+3.8}$). On EOC-Bench, Future reasoning benefits most ($\mathbf{+3.2}$). On EgoOrientBench, the Choose task shows a substantial improvement ($\mathbf{+5.5}$). These non-uniform, category-specific gains beyond SFT indicate that the improvements stem from the proposed plan-then-verify framework and its dense reward design rather than general distillation.

\subsection{Ablation Studies}
\paragraph{RQ1: Is teacher-guided initialization necessary?}
Table~\ref{tab:warmup_ablation} isolates the effect of the 200-step teacher-guided initialization during the RL stage. The model achieves substantial gains (average $+2.9$) even without teacher supervision during RL. Since the SFT policy is initialized using human-annotated data, it already provides a reasonable starting point for self-ranking optimization. These results suggest that the improvements are primarily driven by the reward formulation rather than reliance on teacher guidance.

\paragraph{RQ2: Are ACMG and confidence rewards complementary?} 
Table~\ref{tab:reward_ablation} examines the individual and combined effects of ACMG and the confidence reward. ACMG primarily improves performance on tasks requiring temporal grounding (e.g., EgoPlan $+0.6$), suggesting that aligning predicted plans with future visual observations benefits procedural reasoning. In contrast, the confidence reward yields gains on tasks emphasizing verification consistency (e.g., EOC-Bench $+1.3$), indicating its role in refining third-person audits of the egocentric plan. When combined, the two rewards produce larger improvements (e.g., EOC-Bench $+1.9$) than either component alone, suggesting that temporal grounding and verification-based preference signals provide complementary benefits.

\input{tables/teacher_guidance_reward_ablation}
\input{tables/tags_present_s}

\paragraph{RQ3: Is disentangled reasoning into \protect\egoplan and \protect\exoverify necessary?} Table~\ref{tab:disentangle_ablation} shows that removing either component degrades performance. Eliminating \egoplan significantly degrades egocentric performance (e.g., $-1.8$ on EgoPlan), indicating the importance of modeling action sequences before answer generation. Conversely, removing \exoverify reduces exocentric performance (e.g., $-1.7$ on Video-MME), suggesting that the verification stage contributes to maintaining general third-person reasoning capabilities. Together, these results indicate that the planning and verification components play complementary roles: \egoplan supports task-specific procedural reasoning, while \exoverify helps retain broader cross-view generalization. The full \modelname{} framework balances these components, achieving improved performance across both egocentric and exocentric benchmarks (e.g., $+1.7$ on Video-MME). 

\input{figures/qual}

\paragraph{RQ4: Why does anticipatory grounding (ACMG) improve procedural reasoning?}
Table~\ref{tab:anticipatory_ablation} contrasts present-frame grounding with anticipatory grounding. Models trained with present-frame alignment primarily capture the current static scene, yielding limited improvements (e.g., $+1.0$ on EgoPlan). In contrast, ACMG aligns anticipated actions with future observations over a temporal horizon, encouraging consistency between planned actions and their subsequent outcomes. This anticipatory grounding is associated with larger gains (e.g., $+2.7$ on EgoPlan and $+7.7$ on EgoBlind). As illustrated in Figure~\ref{fig:qualitative} (Left), ACMG correctly anticipates the ``sweep cane'' action before the blind person reaches the curb.

\section{Conclusion}

We presented \modelname{}, a framework for egocentric video reasoning that decomposes reasoning into a plan-then-verify process consisting of egocentric planning and exocentric verification. By coupling this structure with dense rewards for anticipatory grounding and third-person verification, \modelname{} learns to generate procedurally coherent and visually grounded plans and verifications. Empirical results demonstrate improvements in egocentric reasoning while preserving performance on exocentric video understanding tasks, highlighting the effectiveness of cross-perspective reasoning without requiring paired ego-exo supervision. The ability to maintain exocentric performance while specializing in egocentric reasoning opens the possibility of a single unified model serving both first-person assistive applications and general video understanding. 

Future work will extend this framework to streaming egocentric scenarios, where the model must plan and verify incrementally as new frames arrive rather than reasoning over an entire video. Adaptive temporal windows that expand for long, multi-step activities and contract for rapid actions could improve grounding accuracy across diverse tasks. Finally, iterative plan-verify reasoning could enable predicted plans to be refined through repeated cross-perspective feedback when visual evidence is ambiguous or multiple plausible future plans exist.

%% file: tables/main_ego.tex
\begin{table*}[!t]
\centering
\caption{\textbf{Evaluation of \modelname{} on egocentric and exocentric benchmarks.} Best scores are in \textbf{bold}, and improvements from \modelname{} are highlighted in \textcolor{darkgreen}{green} with 95\% confidence interval (CI) margins ($\pm$) via  bootstrap. $^{\ddagger}$ Improvement not statistically significant at the 0.05 level. $^{\diamondsuit}$ Self-trained variant: SFT data and teacher confidence signals generated by Qwen3-VL-8B itself (no external 72B teacher).}
\label{tab:ego_benchmarks}
\begin{adjustbox}{width=\textwidth,center}
\renewcommand{\arraystretch}{1.2}
\fontsize{10pt}{10pt}\selectfont
\setlength{\tabcolsep}{1mm}
\begin{tabular}{lccccc|cccc}
\toprule
& \multicolumn{5}{c}{\textbf{Egocentric Video Understanding}} & \multicolumn{4}{c}{\textbf{Exocentric Video Understanding}} \\
\cmidrule(lr){2-6}\cmidrule(lr){7-10}
\textbf{Model} &
\textbf{EgoBlind} & \textbf{EgoPlan} & \textbf{EgoThink} & \textbf{EOC-Bench} & \textbf{EgoOrient} &
\textbf{MVBench} & \textbf{Video-MME} & \textbf{LVBench} & \textbf{Tomato} \\
\midrule
\rowcolor{gray!15}\multicolumn{10}{c}{\textit{InternVL-3.5 Family}} \\
InternVL-3.5      & 47.8 & 34.0 & 58.5 & 37.1 & 36.3 & 71.4 & 65.6  & 44.4   & 31.4 \\
SFT              & 49.2 & 34.8 & 59.3 & 38.1 & 37.8 & 68.2 & 62.1 & 41.2 & 29.8 \\
GRPO (Format + Answer)    & 49.8 & 35.1 & 59.6 & 38.6 & 38.4 & 68.9 & 63.2 & 41.8 & 30.1 \\
\rowcolor{PastaYellow}
\textbf{\modelname{}}   & 
\textbf{51.4} $\scriptstyle\textcolor{darkgreen}{\textbf{(+3.6}\pm\textbf{0.5)}}$ &
\textbf{35.9} $\scriptstyle\textcolor{darkgreen}{\textbf{(+1.9}\pm\textbf{0.5)}}$ &
\textbf{60.8} $\scriptstyle\textcolor{darkgreen}{\textbf{(+2.3}\pm\textbf{0.5)}}$ &
\textbf{39.7} $\scriptstyle\textcolor{darkgreen}{\textbf{(+2.6}\pm\textbf{0.6)}}$ &
\textbf{40.1} $\scriptstyle\textcolor{darkgreen}{\textbf{(+3.8}\pm\textbf{0.7)}}$ &
\textbf{73.8} $\scriptstyle\textcolor{darkgreen}{\textbf{(+2.4}\pm\textbf{0.6)}}$ &
\textbf{68.3} $\scriptstyle\textcolor{darkgreen}{\textbf{(+2.7}\pm\textbf{0.6)}}$ &
\textbf{47.1} $\scriptstyle\textcolor{darkgreen}{\textbf{(+2.7}\pm\textbf{0.7)}}$ &
\textbf{33.2} $\scriptstyle\textcolor{darkgreen}{\textbf{(+1.8}\pm\textbf{0.6)}}$ \\
\midrule
\rowcolor{gray!15}\multicolumn{10}{c}{\textit{Qwen2.5-VL Family}} \\
Qwen2.5-VL        & 29.7 & 30.2 & 48.2 & 41.6 & 47.6 & 65.2 & 62.2 & 42.8 & 29.3 \\
SFT              & 34.8 & 31.3 & 49.1 & 41.8 & 48.1 & 63.2 & 60.1 & 40.2 & 29.0 \\
GRPO (Format + Answer)    & 35.6 & 31.4 & 49.0 & 42.2 & 48.6 & 66.4 & 62.9 & 40.9 & 29.8 \\
\rowcolor{PastaYellow}
\textbf{\modelname{}}  &
\textbf{37.4} $\scriptstyle\textcolor{darkgreen}{\textbf{(+7.7}\pm\textbf{0.9)}}$ &
\textbf{32.9} $\scriptstyle\textcolor{darkgreen}{\textbf{(+2.7}\pm\textbf{0.6)}}$ &
\textbf{51.9} $\scriptstyle\textcolor{darkgreen}{\textbf{(+3.7}\pm\textbf{0.7)}}$ &
\textbf{44.1} $\scriptstyle\textcolor{darkgreen}{\textbf{(+2.5}\pm\textbf{0.6)}}$ &
\textbf{52.0} $\scriptstyle\textcolor{darkgreen}{\textbf{(+4.4}\pm\textbf{0.8)}}$ &
\textbf{66.8} $\scriptstyle\textcolor{darkgreen}{\textbf{(+1.6}\pm\textbf{0.5)}}$ &
\textbf{64.6} $\scriptstyle\textcolor{darkgreen}{\textbf{(+2.4}\pm\textbf{0.6)}}$ &
\textbf{43.6} $\scriptstyle\textcolor{darkgreen}{\textbf{(+0.8}\pm\textbf{0.5)}}$ &
\textbf{30.2} $\scriptstyle\textcolor{darkgreen}{\textbf{(+0.9}\pm\textbf{0.5)}}$ \\
\midrule
\rowcolor{gray!15}\multicolumn{10}{c}{\textit{Qwen3-VL Family}} \\
Qwen3-VL         & 48.4 & 33.7 & 62.7 & 46.8 & 60.8 & 68.4 & 71.4 & 56.1 & 34.0 \\
SFT             & 50.1 & 34.2 & 63.1 & 47.5 & 61.8 & 68.2 & 71.2 & 56.3 & 34.2 \\
GRPO (Format + Answer)   & 50.8 & 34.5 & 63.4 & 47.9 & 62.3 & 68.1 & 71.1 & 56.4 & 34.3 \\
\rowcolor{PastaYellow}
\textbf{\modelname{}}  &
\textbf{51.9} $\scriptstyle\textcolor{darkgreen}{\textbf{(+3.5}\pm\textbf{0.7)}}$ &
\textbf{35.0} $\scriptstyle\textcolor{darkgreen}{\textbf{(+1.3}\pm\textbf{0.5)}}$ &
\textbf{63.9} $\scriptstyle\textcolor{darkgreen}{\textbf{(+1.2}\pm\textbf{0.5)}}$ &
\textbf{48.6} $\scriptstyle\textcolor{darkgreen}{\textbf{(+1.8}\pm\textbf{0.6)}}$ &
\textbf{63.1} $\scriptstyle\textcolor{darkgreen}{\textbf{(+2.3}\pm\textbf{0.6)}}$ &
\textbf{69.2} $\scriptstyle\textcolor{darkgreen}{\textbf{(+0.8}\pm\textbf{0.5)}}$ &
\textbf{72.2} $\scriptstyle\textcolor{darkgreen}{\textbf{(+0.8}\pm\textbf{0.5)}}$ &
\textbf{56.5} $\scriptstyle\textcolor{gray}{\textbf{(+0.4}\pm\textbf{0.4)}}^{\ddagger}$ &
\textbf{34.5} $\scriptstyle\textcolor{darkgreen}{\textbf{(+0.5}\pm\textbf{0.4)}}$ \\
\midrule
\rowcolor{gray!15}\multicolumn{10}{c}{\textit{Self-Trained}} \\
\rowcolor{cyan!8}
\textbf{\modelname{}}$^{\diamondsuit}$  &
50.2 $\scriptstyle\textcolor{darkgreen}{\textbf{(+1.8)}}$ &
34.1 $\scriptstyle\textcolor{darkgreen}{\textbf{(+0.4)}}$ &
63.0 $\scriptstyle\textcolor{darkgreen}{\textbf{(+0.3)}}$ &
47.4 $\scriptstyle\textcolor{darkgreen}{\textbf{(+0.6)}}$ &
61.5 $\scriptstyle\textcolor{darkgreen}{\textbf{(+0.7)}}$ &
68.3 $\scriptstyle\textcolor{gray}{\textbf{(-0.1)}}$ &
71.5 $\scriptstyle\textcolor{darkgreen}{\textbf{(+0.1)}}$ &
56.0 $\scriptstyle\textcolor{gray}{\textbf{(-0.1)}}$ &
33.9 $\scriptstyle\textcolor{gray}{\textbf{(-0.1)}}$ \\
\bottomrule
\end{tabular}
\end{adjustbox}
\end{table*}

%% file: tables/qwen2vl_based.tex
\begin{table*}[t!]
\centering
\caption{\textbf{Comparison with State of the Art.} All results are reproduced by us.}
\label{tab:qwen2vl_based}
\begin{adjustbox}{width=\columnwidth,center}
\renewcommand{\arraystretch}{1.2}
\fontsize{8pt}{9pt}\selectfont
\setlength{\tabcolsep}{1mm}
\begin{tabular}{lcccc|cccc}
\toprule
& \multicolumn{4}{c}{\textbf{Egocentric Video Reasoning}} & \multicolumn{4}{c}{\textbf{Basic Exocentric Understanding}} \\
\cmidrule(lr){2-5}\cmidrule(lr){6-9}
\textbf{Variant} & \textbf{EgoBlind} & \textbf{EgoPlan} & \textbf{EOC-Bench} & \textbf{EgoSchema} & \textbf{MMMU} & \textbf{DocVQA} & \textbf{MME (Perception)} & \textbf{VQAv2}\\
\midrule
Qwen2-VL & 41.0 & 32.5 & 39.6 & 64.2 & 51.0 & 79.5 & 1677 & 80.4 \\
SFT & 44.2 & 33.1 & 40.8 & 65.4 & 49.8 & 78.1 & 1654 & 79.2 \\
GRPO (Format + Answer) & 45.6 & 33.6 & 41.4 & 66.1 & 50.4 & 78.9 & 1665 & 79.8 \\
\rowcolor{gray!10}
EgoThinker~\cite{egothinker} & 43.8 & 34.1 & 37.1 & \textbf{68.2} & 
49.2 $\scriptstyle\textcolor{darkred}{\textbf{(-1.8)}}$ & 
69.0 $\scriptstyle\textcolor{darkred}{\textbf{(-10.5)}}$ & 
1644 $\scriptstyle\textcolor{darkred}{\textbf{(-33)}}$ & 
78.1 $\scriptstyle\textcolor{darkred}{\textbf{(-2.3)}}$ \\
\rowcolor{PastaYellow}
\textbf{EgoVITA} & 
\textbf{46.8} $\scriptstyle\textcolor{darkgreen}{\textbf{(+5.8}\pm\textbf{0.3)}}$ & 
\textbf{34.6} $\scriptstyle\textcolor{darkgreen}{\textbf{(+2.1}\pm\textbf{0.4)}}$ & 
\textbf{42.7} $\scriptstyle\textcolor{darkgreen}{\textbf{(+3.1}\pm\textbf{0.2)}}$ & 
67.9 $\scriptstyle\textcolor{darkgreen}{\textbf{(+3.7}\pm\textbf{0.5)}}$ & 
\textbf{51.3} $\scriptstyle\textcolor{darkgreen}{\textbf{(+0.3}\pm\textbf{0.4)}}$ & 
\textbf{80.1} $\scriptstyle\textcolor{darkgreen}{\textbf{(+0.6}\pm\textbf{0.3)}}$ & 
\textbf{1682} $\scriptstyle\textcolor{darkgreen}{\textbf{(+5}\pm\textbf{7)}}$ & 
\textbf{81.2} $\scriptstyle\textcolor{darkgreen}{\textbf{(+0.8}\pm\textbf{0.6)}}$ \\
\bottomrule
\end{tabular}
\end{adjustbox}
\end{table*}

%% file: figures/attention.tex
\begin{figure*}[!t]
\centering
\includegraphics[width=\textwidth]{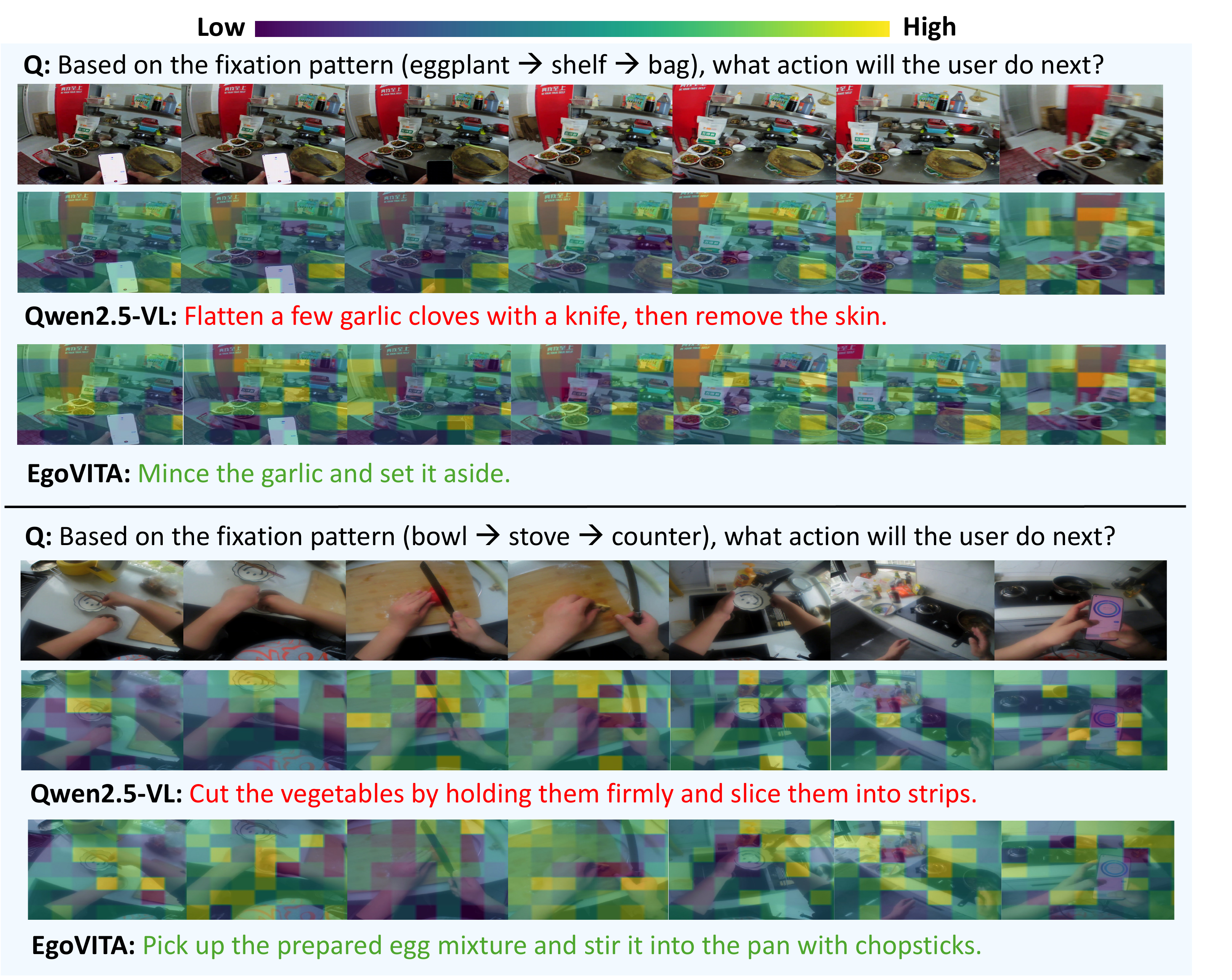}
\caption{\textbf{\modelname{} learns to ``look ahead'' in egocentric video.} EAGLE~\cite{eagle} visualizations show that \modelname{} progressively concentrates attention on task-relevant objects (e.g., garlic, egg bowl, pan) and hand regions across frames, whereas Qwen2.5-VL shows scattered, temporally static activations over background elements (e.g., shelves, walls). The focused, temporally evolving attention patterns in \modelname{} are associated with correct action predictions (\textcolor{darkgreen}{green}), whereas the baseline often produces incorrect or visually inconsistent outputs (\textcolor{red}{red}).}
\label{fig:attention}
\end{figure*}

%% file: figures/tsne_noun.tex
\begin{figure}[!t]
    \centering
    \begin{minipage}[t]{0.48\columnwidth}
        \centering
        \includegraphics[width=\linewidth]{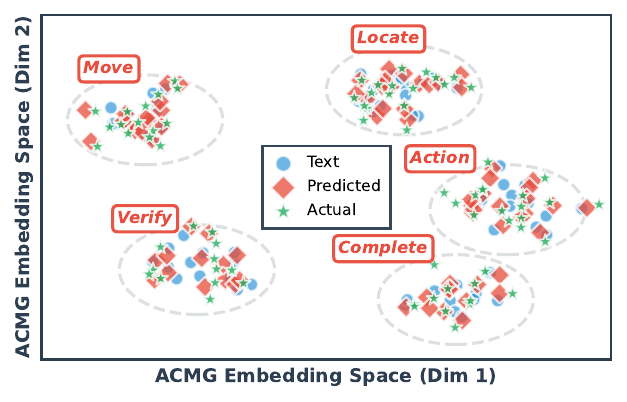}
        \caption{\textbf{ACMG Embedding Structure.} t-SNE of text (blue), predicted (red), and actual (green) embeddings. Alignment within distinct action clusters indicates cross-modal grounding.}
        \label{fig:acmg_tsne}
    \end{minipage}
    \hfill
    \begin{minipage}[t]{0.48\columnwidth}
        \centering
        \includegraphics[width=\linewidth]{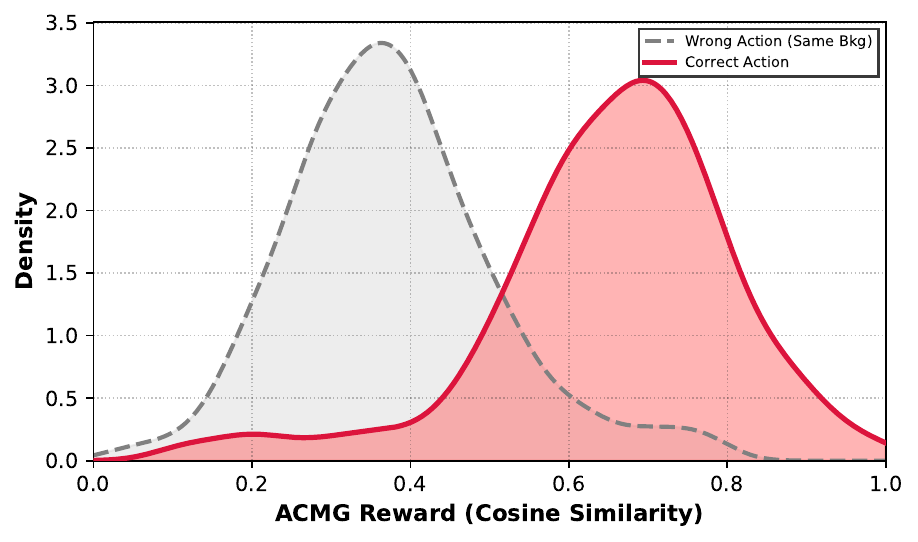}
        \caption{\textbf{Static Bias Check.}
        Reward density for matched vs.\ same-scene/wrong-action pairs. The gap ($\Delta=0.33$) shows sensitivity to temporal dynamics rather than static backgrounds}
        \label{fig:acmg_bias}
    \end{minipage}
\end{figure}

%% file: tables/category_breakdown.tex
\begin{table}[t!]
\centering
\caption{\textbf{Task-level breakdown.}
EgoBlind categories from~\cite{xiao2025egoblind} are Tool Use (TU), Information Reading (IR), Navigation (NV), Safety Warning (SW), Social Communication (SC), and Other Resources (OR).
EOC-Bench temporal dimensions from~\cite{yuan2025eoc} are Past (Pst), Present (Prs), and Future (Fut).
EgoOrientBench tasks from~\cite{jung2025is} are Choose (Cho), Verify (Ver), and Freeform (FF).}
\label{tab:category_breakdown}
\begin{adjustbox}{width=\columnwidth,center}
\renewcommand{\arraystretch}{1.1}
\fontsize{6pt}{7pt}\selectfont
\setlength{\tabcolsep}{0.9mm}
\begin{tabular}{l|ccccccc|cccc|cccc}
\toprule
& \multicolumn{7}{c|}{\textbf{EgoBlind}} & \multicolumn{4}{c|}{\textbf{EOC-Bench}} & \multicolumn{4}{c}{\textbf{EgoOrientBench}} \\
\cmidrule(lr){2-8}\cmidrule(lr){9-12}\cmidrule(lr){13-16}
\textbf{Method} & \textbf{TU} & \textbf{IR} & \textbf{NV} & \textbf{SW} & \textbf{SC} & \textbf{OR} & \textbf{Avg} & \textbf{Pst} & \textbf{Prs} & \textbf{Fut} & \textbf{Avg} & \textbf{Cho} & \textbf{Ver} & \textbf{FF} & \textbf{Avg} \\
\midrule
Qwen2.5-VL & 34.3 & 35.0 & 12.8 & 30.8 & 37.8 & 27.5 & 29.7 & 33.7 & 48.8 & 42.2 & 41.6 & 45.0 & 62.0 & 35.8 & 47.6 \\
SFT        & 36.3 & 38.6 & 15.8 & 41.2 & 41.4 & 35.5 & 34.8 & 33.9 & 49.1 & 42.4 & 41.8 & 45.5 & 62.5 & 36.3 & 48.1 \\
GRPO       & 36.7 & 39.2 & 16.3 & 42.9 & 41.9 & 36.6 & 35.6 & 34.2 & 49.4 & 43.0 & 42.2 & 46.0 & 63.0 & 36.8 & 48.6 \\
\rowcolor{PastaYellow}
\textbf{\modelname{}} & \textbf{37.4} & \textbf{40.5} & \textbf{17.4} & \textbf{46.6} & \textbf{43.2} & \textbf{39.3} & \textbf{37.4} $\scriptstyle\textcolor{darkgreen}{\textbf{(+7.7)}}$ & \textbf{36.5} & \textbf{50.2} & \textbf{45.6} & \textbf{44.1} $\scriptstyle\textcolor{darkgreen}{\textbf{(+2.5)}}$ & \textbf{51.0} & \textbf{65.5} & \textbf{39.5} & \textbf{52.0} $\scriptstyle\textcolor{darkgreen}{\textbf{(+4.4)}}$ \\
\bottomrule
\end{tabular}
\end{adjustbox}
\end{table}

%% file: tables/teacher_guidance_reward_ablation.tex
\begin{table}[t!]
\centering

\begin{minipage}[t]{0.48\columnwidth}
    \centering
    \caption{\textbf{Teacher-Guided Warm-up.}
    Effect of removing the 200-step teacher phase from the confidence reward.}
    \label{tab:warmup_ablation}

    \begin{adjustbox}{width=\linewidth,center}
    \renewcommand{\arraystretch}{1.2}
    \fontsize{8pt}{9pt}\selectfont
    \setlength{\tabcolsep}{1mm}
    \begin{tabular}{l cccc}
        \toprule
        & \multicolumn{4}{c}{\textbf{Egocentric Benchmarks}} \\
        \cmidrule(lr){2-5}
        \textbf{Variant} & \textbf{EgoBlind} & \textbf{EgoPlan} & \textbf{EgoThink} & \textbf{EOC-Bench} \\
        \midrule
        Qwen2.5-VL-7B & 29.7 & 30.2 & 48.2 & 41.6 \\
        \midrule
        No Teacher Warm-up &
        36.0 $\scriptstyle\textcolor{darkgreen}{\textbf{(+6.3)}}$ &
        31.8 $\scriptstyle\textcolor{darkgreen}{\textbf{(+1.6)}}$ &
        50.5 $\scriptstyle\textcolor{darkgreen}{\textbf{(+2.3)}}$ &
        43.0 $\scriptstyle\textcolor{darkgreen}{\textbf{(+1.4)}}$ \\
        \rowcolor{PastaYellow}
        \textbf{\modelname{} (Full)} &
        \textbf{37.4} $\scriptstyle\textcolor{darkgreen}{\textbf{(+7.7)}}$ &
        \textbf{32.9} $\scriptstyle\textcolor{darkgreen}{\textbf{(+2.7)}}$ &
        \textbf{51.9} $\scriptstyle\textcolor{darkgreen}{\textbf{(+3.7)}}$ &
        \textbf{44.1} $\scriptstyle\textcolor{darkgreen}{\textbf{(+2.5)}}$ \\
        \bottomrule
    \end{tabular}
    \end{adjustbox}
\end{minipage}
\hfill
\begin{minipage}[t]{0.48\columnwidth}
    \centering
    \caption{\textbf{ACMG and confidence reward on Qwen2.5-VL.}
    Removing either component degrades performance.}
    \label{tab:reward_ablation}

    \begin{adjustbox}{width=\linewidth,center}
    \renewcommand{\arraystretch}{1.2}
    \fontsize{8pt}{9pt}\selectfont
    \setlength{\tabcolsep}{1mm}
    \begin{tabular}{lcccc}
        \toprule
        \textbf{Variant} & \textbf{EgoPlan} & \textbf{EOC-Bench} & \textbf{Video-MME} & \textbf{Tomato} \\
        \midrule
        GRPO (Format + Answer) & 31.4 & 42.2 & 62.9 & 29.8 \\
        \rowcolor{gray!10}
        + ACMG only & 32.0 & 43.1 & 63.4 & 29.9 \\
        + Confidence only & 32.3 & 43.5 & 63.8 & 30.0 \\
        \rowcolor{PastaYellow}
        \textbf{\modelname{} (Full)} &
        \textbf{32.9} $\scriptstyle\textcolor{darkgreen}{\textbf{(+1.5)}}$ &
        \textbf{44.1} $\scriptstyle\textcolor{darkgreen}{\textbf{(+1.9)}}$ &
        \textbf{64.6} $\scriptstyle\textcolor{darkgreen}{\textbf{(+1.7)}}$ &
        \textbf{30.2} $\scriptstyle\textcolor{darkgreen}{\textbf{(+0.4)}}$ \\
        \bottomrule
    \end{tabular}
    \end{adjustbox}
\end{minipage}

\end{table}

%% file: tables/tags_present_s.tex
\begin{table}[t!]
\centering

\begin{minipage}[t]{0.48\columnwidth}
    \centering
    \caption{\textbf{Disentangled reasoning.}
    Removing \protect\egoplan harms egocentric reasoning, while removing \protect\exoverify reduces exocentric performance.}
    \label{tab:disentangle_ablation}
    \begin{adjustbox}{width=\linewidth,center}
    \renewcommand{\arraystretch}{1.2}
    \fontsize{8pt}{9pt}\selectfont
    \setlength{\tabcolsep}{1mm}
    \begin{tabular}{lcccc}
        \toprule
        \textbf{Variant} & \textbf{EgoPlan} & \textbf{EOC-Bench} & \textbf{Video-MME} & \textbf{Tomato} \\
        \midrule
        GRPO (Format + Answer) & 31.4 & 42.2 & 62.9 & 29.8 \\
        w/o \protect\egoplan & 29.6 & 40.1 & 63.0 & 29.7 \\
        w/o \protect\exoverify & 31.5 & 42.5 & 61.2 & 28.3 \\
        \rowcolor{PastaYellow}
        \textbf{\modelname{} (Full)} &
        \textbf{32.9} $\scriptstyle\textcolor{darkgreen}{\textbf{(+1.5)}}$ &
        \textbf{44.1} $\scriptstyle\textcolor{darkgreen}{\textbf{(+1.9)}}$ &
        \textbf{64.6} $\scriptstyle\textcolor{darkgreen}{\textbf{(+1.7)}}$ &
        \textbf{30.2} $\scriptstyle\textcolor{darkgreen}{\textbf{(+0.4)}}$ \\
        \bottomrule
    \end{tabular}
    \end{adjustbox}
\end{minipage}
\hfill
\begin{minipage}[t]{0.48\columnwidth}
    \centering
    \caption{\textbf{Anticipatory (ACMG) vs. Present Grounding.} Comparison with a present-frame baseline (grounding on frame $t$). Both improve performance, with ACMG yielding substantially larger gains.}
    \label{tab:anticipatory_ablation}
    \begin{adjustbox}{width=\linewidth,center}
    \renewcommand{\arraystretch}{1.2}
    \fontsize{6pt}{7pt}\selectfont
    \setlength{\tabcolsep}{1mm}
    \begin{tabular}{lcccc}
        \toprule
        \textbf{Model} & \textbf{EgoBlind} & \textbf{EgoPlan} & \textbf{EgoThink} & \textbf{EOC-Bench} \\
        \midrule
        Qwen2.5-VL-7B & 29.7 & 30.2 & 48.2 & 41.6 \\
        \textbf{Present}  &
        33.8 $\scriptstyle\textcolor{darkgreen}{\textbf{(+4.1)}}$ &
        31.2 $\scriptstyle\textcolor{darkgreen}{\textbf{(+1.0)}}$ &
        50.1 $\scriptstyle\textcolor{darkgreen}{\textbf{(+1.9)}}$ &
        42.8 $\scriptstyle\textcolor{darkgreen}{\textbf{(+1.2)}}$ \\
        \rowcolor{PastaYellow}
        \textbf{ACMG} &
        \textbf{37.4} $\scriptstyle\textcolor{darkgreen}{\textbf{(+7.7)}}$ &
        \textbf{32.9} $\scriptstyle\textcolor{darkgreen}{\textbf{(+2.7)}}$ &
        \textbf{51.9} $\scriptstyle\textcolor{darkgreen}{\textbf{(+3.7)}}$ &
        \textbf{44.1} $\scriptstyle\textcolor{darkgreen}{\textbf{(+2.5)}}$ \\
        \bottomrule
    \end{tabular}
    \end{adjustbox}
\end{minipage}
\label{tab:anitca}
\end{table}

%% file: figures/qual.tex
\begin{figure*}[!t]
\centering
\includegraphics[width=\textwidth]{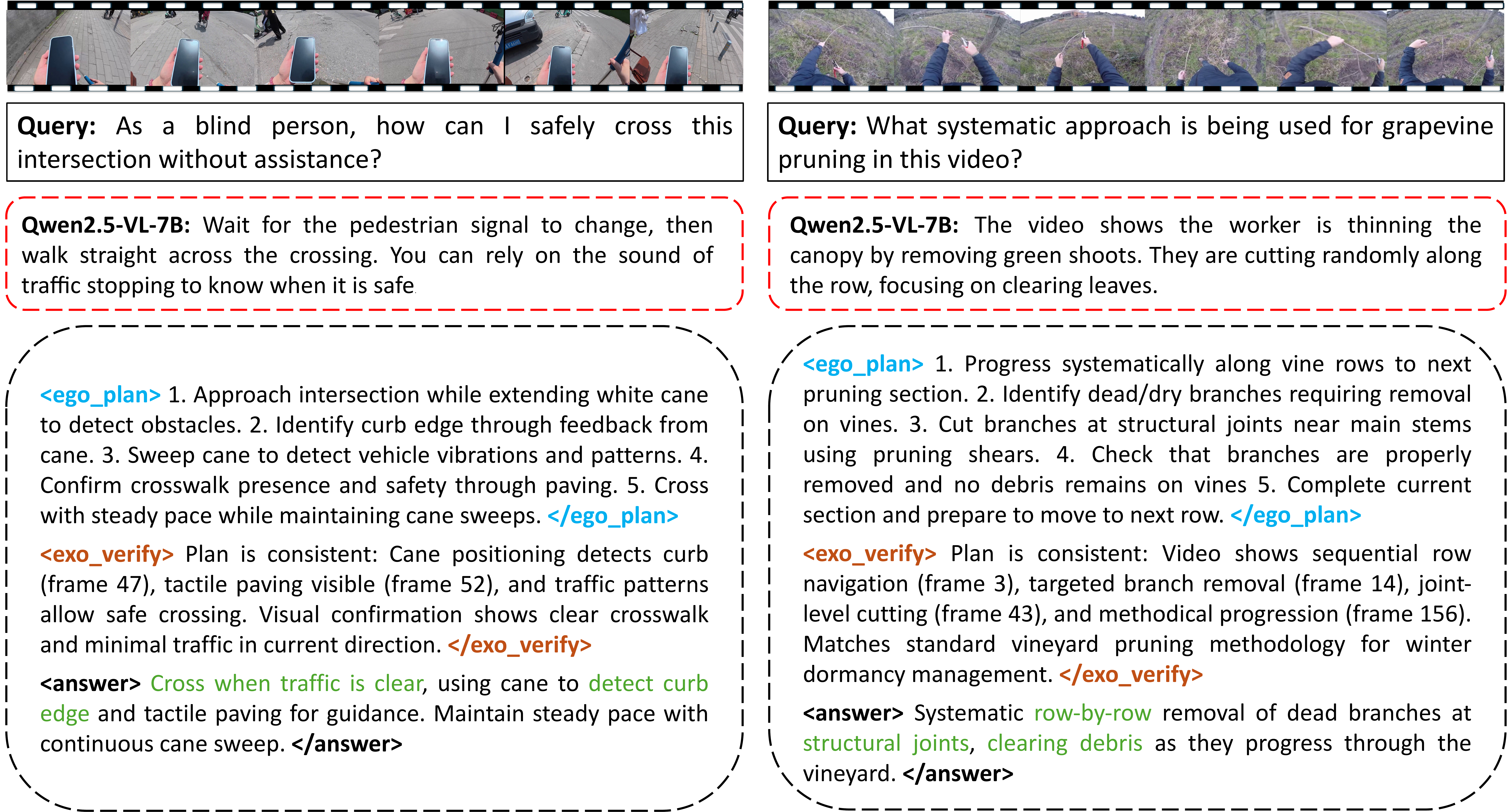}
\caption{\textbf{\modelname{} Qualitative Examples.} \textbf{Left:} For a blind person, the \protect\egoplan generates sequential, task-oriented actions. \textbf{Right:} For a procedural task, \protect\egoplan outlines a structured step-by-step plan. In both cases, \protect\exoverify  evaluates consistency with relevant visual frames, producing a grounded \protect\answer. In contrast, Qwen2.5-VL produces \textit{unsafe} visually inconsistent responses.}
\label{fig:qualitative}
\end{figure*}

%% file: sec/appendix.tex
\clearpage
  \setcounter{section}{0}
  \setcounter{figure}{0}
  \setcounter{table}{0}
\setcounter{page}{1}
\title{EgoVITA: Learning to Plan and Verify for Egocentric Video Reasoning}

\begin{center}
  \Large\textbf{EgoVITA: Learning to Plan and Verify for Egocentric Video Reasoning}\\[3mm]
  \large\textbf{Supplementary Material}
  \end{center}
  \vspace{4mm}

\authorrunning{Y.~Kulkarni and P.~Fazli}

\section{Ablation Studies}
 \thispagestyle{empty}
\input{tables/mean_max}
\paragraph{RQ1: Why Max Instead of Average Pooling for $R_{\mathrm{ACMG}}$?}

A valid concern is that \textit{max} pooling (Eq. 8) could reward spurious, high-similarity outliers, whereas a \textit{mean} pooling operation might better capture durational events (e.g., ``walk to the bus stop''). In Table \ref{tab:pooling_ablation}, we directly ablate this by replacing \textit{max} pooling with \textit{mean} pooling across the $N=16$ future frames. The results show that \textit{mean} pooling is significantly less effective, providing only a fraction of the performance gain (e.g., $+3.8$ vs. $+7.7$ on EgoBlind, and $+1.9$ vs. $+3.7$ on EgoThink).

The \textit{mean} operation's weakness is that it \textbf{dilutes the reward signal}. A correct durational clause like ``walk to the bus stop'' will have high similarity with \textit{some}, but not necessarily \textit{all}, of the $N=16$ future frames. Averaging across the entire window (including the non-matching frames) punishes this correct, but sparse, temporal alignment, resulting in a weak, non-descriptive reward. In contrast, \texttt{max} pooling is ``robust to temporal shifts'' because it finds the \textbf{peak evidence} that the predicted event \textit{did} occur within the window, regardless of its exact timing or duration. This provides a much stronger and more stable learning signal for grounding, and the empirical results suggest this benefit outweighs the risk of spurious outliers.

\paragraph{RQ2: Impact of Exocentric Regularization.}
\input{figures/exo_vs_steps}

Figure \ref{fig:exo_regularization_ablation} demonstrates the importance of exocentric regularization for preventing catastrophic forgetting. Training on egocentric data via SFT causes an initial performance drop on exocentric benchmarks, as seen by the `RL Step 0’ (SFT) performance being below the `Base Model’ across all four tasks. This drop occurs because SFT overfits to the egocentric data distribution. Without intervention, this forgetting continues, as shown by the `w/o Regularization’, where performance progressively degrades. In contrast, \modelname{}’s regularization, which interleaves GRPO updates with training on a held-out exocentric dataset, reverses this trend. The model not only recovers from the initial SFT drop but ultimately improves upon the original base model's performance. This confirms that our approach enables the model to specialize in egocentric reasoning while preserving and even enhancing its general exocentric understanding, tackling a key challenge in ego-exo knowledge transfer.

\paragraph{RQ3: Does exocentric regularization help egocentric reasoning?}
Figure~\ref{fig:exo_regularization_ego_perfomance_ablation} shows that moderate exocentric regularization ($\lambda_{\mathrm{exo}}{=}0.05$) improves egocentric performance across all four benchmarks, peaking at $32.9$ on EgoPlan and $51.9$ on EgoThink. This is because a small exocentric signal prevents the model from overfitting to the narrow egocentric training distribution, preserving the general visual reasoning capabilities that also benefit first-person understanding. However, stronger regularization ($\lambda_{\mathrm{exo}} \geq 0.2$) shifts the optimization toward exocentric objectives, diluting the egocentric reward signal and causing sharp performance drops. At $\lambda_{\mathrm{exo}}{=}0.5$, performance approaches the GRPO baseline on most tasks, indicating that excessive regularization effectively neutralizes the dense reward training. 

\input{figures/ego_vs_regularization}

\paragraph{RQ4: How does training different vision components affect egocentric reasoning?}
Table~\ref{tab:component_ablation} shows that joint adaptation of all vision components is critical for strong performance. Training with a \textit{frozen vision pipeline} yields the smallest gain ($+2.4$ on EgoBlind) as the visual features are static. Adapting only the \textit{projector} improves alignment ($+4.5$), while adapting only the \textit{encoder} yields a larger gain ($+6.1$) but can lead to conflicts between the encoder outputs and the downstream projector. The best performance ($+7.7$) is achieved by jointly training both encoder and projector (\textit{Full}), demonstrating that complete end-to-end visual adaptation is necessary for robust egocentric reasoning.
\input{tables/vision_encoder}

\paragraph{RQ5: Sensitivity to Anticipation Window $N$.}
We analyze the sensitivity of $R_{\mathrm{ACMG}}$ to $N$, the number of future frames used for grounding, in Table \ref{tab:n_frames_ablation}. Performance is clearly sensitive to this hyperparameter. A small window, such as $N=4$ or $N=8$, yields lower gains. This is because the anticipation window is too myopic: a correct plan clause (e.g., ``Grab the bottle'') may not have its corresponding visual outcome (the bottle being grabbed) appear within such a short temporal window, unfairly penalizing valid long-term plans. Conversely, a large window ($N=32$) also degrades performance compared to our chosen value. $R_{\mathrm{ACMG}}$ relies on a $\max$ operation over the $N$ future frames to find the best-aligned visual state. A window that is too large makes this signal noisy, increasing the risk of rewarding a spurious or coincidental visual match that is not causally linked to the \egoplan clause. Therefore, $N=16$ represents an effective balance, providing a flexible enough window to handle temporal shifts in actions while remaining focused enough to reward precise, causal predictions.
\input{figures/acmg_temporal_grounding}
\input{tables/frames_ablation}

\input{figures/acmg_box}

\paragraph{RQ6: How does $R_{\mathrm{ACMG}}$ reward vary across clause positions?} We analyze 500 \egoplan samples from EgoPlan~\cite{chen2023egoplan}, which naturally decompose into five semantic positions: (1) Move (approach), (2) Locate (target identification), (3) Action (perform task), (4) Verify (state confirmation), and (5) Complete (goal achievement). Clauses are tagged into the five action categories using GPT-4o~\cite{hurst2024gpt}. These categories are not rigid but reflect a common semantic structure in procedural tasks. Figure~\ref{fig:acmg_grounding} provides a detailed visualization of $R_{\mathrm{ACMG}}$ grounding for a single clause, while Figure~\ref{fig:clause_position} shows that $R_{\mathrm{ACMG}}$ provides position-sensitive grounding signals.
Middle clauses (positions $2$--$3$: corresponding to target localization and action execution) receive significantly higher 
$R_{\mathrm{ACMG}}$ rewards (mean = $0.68$--$0.71$) and show stronger correlations with final answer correctness ($\rho = 0.52$--$0.67$) compared to navigation (position $1$, $\rho = 0.45$) or completion steps (position $5$, $\rho = 0.41$). 

This variation demonstrates that $R_{\mathrm{ACMG}}$ is adaptively selective, 
providing stronger grounding signals for visually concrete actions while appropriately moderating 
rewards for abstract planning phases. The correlation pattern validates that $R_{\mathrm{ACMG}}$ focuses on task-critical visual grounding, i.e., clauses with higher $R_{\mathrm{ACMG}}$ rewards are indeed more 
predictive of correct final answers, confirming that the reward mechanism aligns with meaningful 
task structure rather than providing noisy uniform feedback.
\input{tables/qwen3vl_teacher}

\paragraph{RQ7: Is the framework sensitive to teacher model choice?}

Table~\ref{tab:qwen3vl_based} demonstrates that \modelname{}'s gains are not dependent on a specific teacher model. When using Qwen3-VL-30B as the teacher instead of Qwen2.5-VL-72B, the framework achieves $\mathbf{+8.8}$ on EgoBlind (compared to $\mathbf{+7.7}$ with the original teacher), confirming that the improvements generalize across teacher models. Moreover,  \modelname{}$^{\diamondsuit}$ in Table~1 of the main paper, which uses no external teacher, still outperforms the base Qwen3-VL-8B on all egocentric benchmarks.

\input{figures/qual_supp}

\section{Qualitative Examples}
We provide a direct comparison between \modelname{} and the Qwen2.5-VL-7B in Figure \ref{fig:qualitative_supp}, illustrating how our two-stage architecture resolves common MLLM failure modes in egocentric reasoning.

In the Kitchen Cleanup example (Left), the baseline fails to ground its reasoning in the visual evidence. It exhibits temporal hallucination by adding a ``cabinet storage'' step that never occurs and spatial misalignment by ignoring the actual sink washing process shown in the video. Consequently, it provides a plausible but incorrect recommendation to ``skip checking cleanliness,'' assuming the dishwasher does everything. \modelname{}, however, generates a grounded plan that prioritizes preconditions. The verification module explicitly cites the ``active dish washing'' in Frames $6$-$8$ and the ``dishwasher space check'' in Frame $5$, correcting the logic to ensure a systematic and visually consistent cleanup procedure.

In the Baking Process example (Right), the baseline suffers from hallucinations, introducing factual errors (mentioning ``eggs'' not present in the scene) and spatial misalignment (claiming a ``whisk'' is used when the video clearly shows a spoon). It further fails in temporal reasoning by suggesting ``immediate baking'' without the necessary intermediate steps. \modelname{} avoids these pitfalls through its ACMG reward. By predicting the visual features of the next steps (e.g., ``Find the appropriate measuring spoon''), the model correctly identifies the fine-grained spoon-flour interaction. The exocentric verification further confirms the ``measuring'' action at Frame $18$, ensuring the response is strictly faithful to the observed video content.

Figure~\ref{fig:exo_fixing} further demonstrates how the exocentric verification stage actively corrects the egocentric plan. In the cell culture example (Left), \exoverify identifies that the plan omits critical sterilization steps (flaming, ethanol use) and a swirling motion needed for even distribution, leading to a final answer that recovers these steps. In the sandwich example (Right), \exoverify flags missing slicing and layering details by cross-referencing specific timestamps, producing a more complete answer. These cases illustrate that the perspective transformation from ego to exo is not prompt role-play; it catches procedurally significant omissions that would otherwise propagate to the final response.

\input{figures/exo_fixing}

\section{Failure Case and Rollout Analysis}
To analyze failure modes, we use Gemini-3.1-Flash to categorize 3,278 Qwen2.5-VL training rollouts into six outcome types (Table~\ref{tab:rollout_analysis}). Types A and B correspond to successful cases, where the plan is either correct from the outset (56\%) or corrected by \exoverify (15\%). Type C captures cases where \exoverify confirms an incorrect plan (10\%), while Type D represents failed correction attempts (6\%). Types E and F isolate limitations of ACMG, where rewards are noisy due to outcomes falling beyond the $N{=}16$ window (10\%) or biased by static scene features (3\%). When \egoplan was incorrect (B+C+D, 993 cases), \exoverify rejected the plan in $69$\% of cases (B+D, 681) and confirmed it in only 31\% (C, 312), indicating that the verifier generally corrects planner errors rather than merely rationalizing them.
Figure~\ref{fig:failure_case} illustrates a combined Type C+F failure. After the person exits the scene, the sink remains visible within the $N{=}16$ temporal window, causing ACMG to reward the incorrect ``cleanup'' action based on persistent object co-occurrence. \exoverify then reinforces the error by confirming the same flawed reasoning.

\input{figures/failure_analysis}

\section{Anticipation Head Architecture} 

\input{tables/arch}

The Anticipation Head Table~\ref{tab:mlp_arch} is a lightweight, two-layer Multilayer Perceptron (MLP) designed to project the LLM's hidden textual states into the shared visual embedding space. It consists of a down-projection linear layer, a GELU activation function, and a final linear layer that maps to the visual encoder's output dimension. We apply Layer Normalization pre-activation to stabilize training. This simple design enables efficient cross-modal alignment without introducing significant computational overhead.

\section{Prompt Templates}
To establish robust supervision for our two-stage RL framework, we generate high-quality training data using Qwen2.5-VL-72B~\cite{qwen2_5vl}. For the egocentric planning stage, we utilize the prompt in Figure \ref{fig:sft-prompt} to transform raw EgoProceL~\cite{egoprocel} temporal annotations into structured, step-by-step egocentric plans. Complementing this, the prompt in Figure \ref{fig:verify-prompt} is applied to the HD-EPIC~\cite{perrett2025hd} dataset to generate third-person ``exocentric verifications,'' which teach the model to verify its plans against observable visual evidence.

\input{figures/ego_plan}
\input{figures/exo_verify}

%% file: tables/mean_max.tex
\begin{table}[htbp!]
\centering
\caption{\textbf{Ablation on ACMG Pooling Strategy.} We compare Max Pooling (used in \modelname{}) against Mean Pooling for the $R_{ACMG}$ reward. Results on Qwen2.5-VL-7B.}
\label{tab:pooling_ablation}
\begin{adjustbox}{width=0.7\columnwidth,center}
\renewcommand{\arraystretch}{1.2}
\fontsize{6pt}{7pt}\selectfont
\setlength{\tabcolsep}{1mm}
\begin{tabular}{l cccc}
\toprule
 & \multicolumn{4}{c}{\textbf{Egocentric Benchmarks}} \\
\cmidrule(lr){2-5}
\textbf{ACMG Variant} & \textbf{EgoBlind} & \textbf{EgoPlan} & \textbf{EgoThink} & \textbf{EOC-Bench} \\
\midrule
Qwen2.5-VL-7B & 29.7 & 30.2 & 48.2 & 41.6 \\
\midrule
$R_{\mathrm{ACMG}}$ (Mean Pooling) & 33.5 $\scriptstyle\textcolor{darkgreen}{\textbf{(+3.8)}}$ & 31.5 $\scriptstyle\textcolor{darkgreen}{\textbf{(+1.3)}}$ & 50.1 $\scriptstyle\textcolor{darkgreen}{\textbf{(+1.9)}}$ & 42.6 $\scriptstyle\textcolor{darkgreen}{\textbf{(+1.0)}}$ \\
\rowcolor{PastaYellow}
\textbf{$R_{ACMG}$ (Max Pooling) (\modelname{})} & 
\textbf{37.4} $\scriptstyle\textcolor{darkgreen}{\textbf{(+7.7)}}$ & 
\textbf{32.9} $\scriptstyle\textcolor{darkgreen}{\textbf{(+2.7)}}$ & 
\textbf{51.9} $\scriptstyle\textcolor{darkgreen}{\textbf{(+3.7)}}$ & 
\textbf{44.1} $\scriptstyle\textcolor{darkgreen}{\textbf{(+2.5)}}$ \\
\bottomrule
\end{tabular}
\end{adjustbox}
\end{table}

%% file: figures/exo_vs_steps.tex
\begin{figure*}[!t]
\centering
\begin{subfigure}[b]{0.33\linewidth}
\centering
\includegraphics[width=\linewidth]{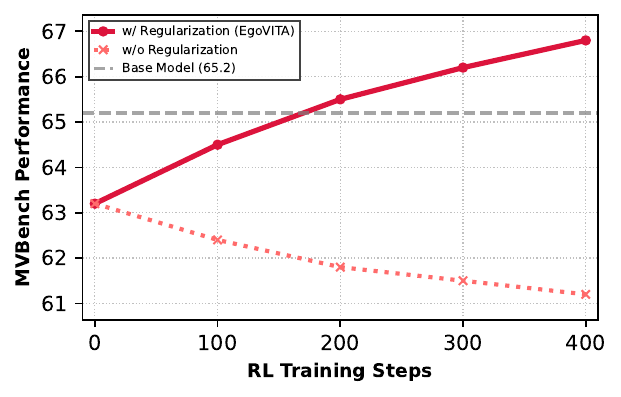}
\caption{MVBench}
\label{fig:mvbench_reg}
\end{subfigure}
\hspace{0.1cm}
\begin{subfigure}[b]{0.33\linewidth}
\centering
\includegraphics[width=\linewidth]{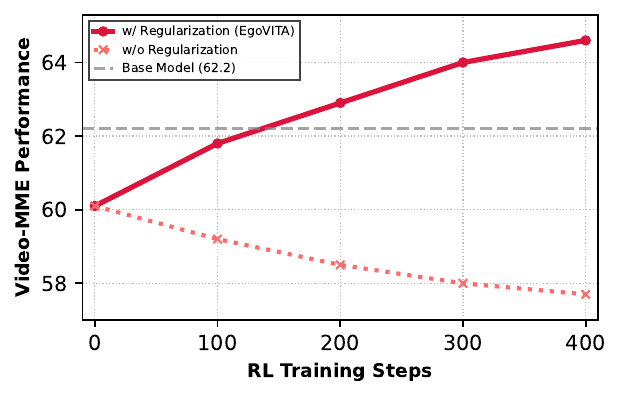}
\caption{Video-MME}
\label{fig:videomme_reg}
\end{subfigure}

\vspace{0.1cm} 

\begin{subfigure}[b]{0.33\linewidth}
\centering
\includegraphics[width=\linewidth]{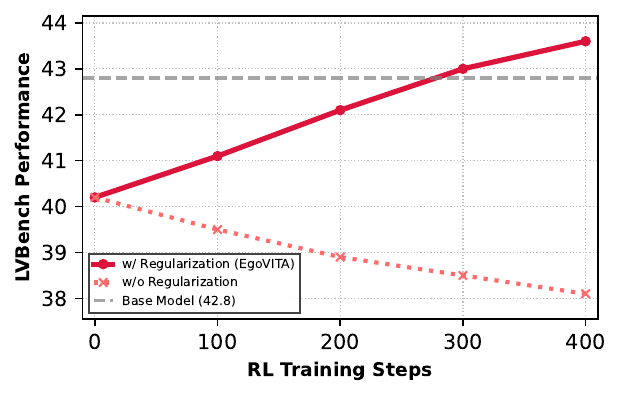}
\caption{LVBench}
\label{fig:lvbench_reg}
\end{subfigure}
\hspace{0.1cm}
\begin{subfigure}[b]{0.33\linewidth}
\centering
\includegraphics[width=\linewidth]{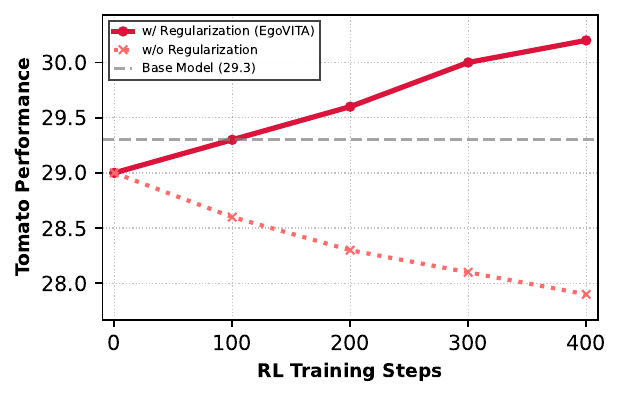}
\caption{TOMATO}
\label{fig:tomato_reg}
\end{subfigure}
\caption{\textbf{Ablation on Exocentric Regularization $\lambda_{\mathrm{exo}}$.} The plots show performance on four exocentric benchmarks (Qwen2.5-VL family). Training on egocentric data via SFT causes an initial performance drop, or catastrophic forgetting. Without regularization, this decline continues. \modelname{}’s exocentric regularization reverses this forgetting, recovering performance and ultimately improving upon the original base model.}
\label{fig:exo_regularization_ablation}
\end{figure*}

%% file: figures/ego_vs_regularization.tex
\begin{figure*}[!t]
\centering
\begin{subfigure}[b]{0.33\linewidth}
\centering
\includegraphics[width=\linewidth]{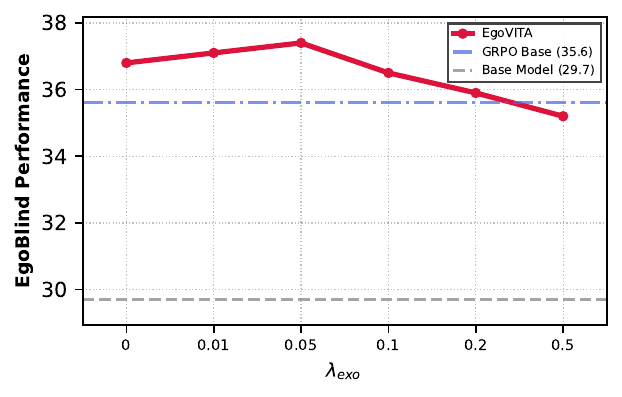}
\caption{EgoBlind}

\end{subfigure}
\hspace{0.1cm}
\begin{subfigure}[b]{0.33\linewidth}
\centering
\includegraphics[width=\linewidth]{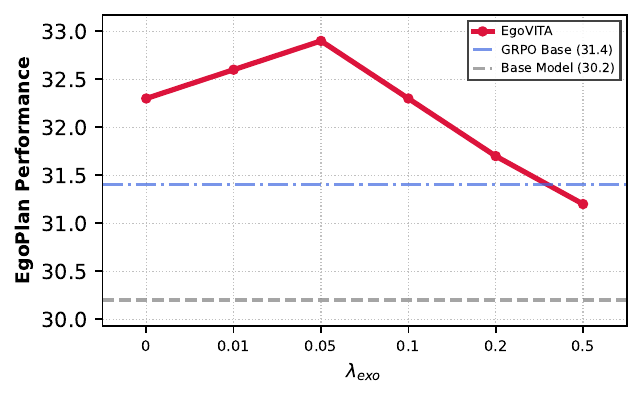}
\caption{EgoPlan}

\end{subfigure}

\vspace{0.1cm} 

\begin{subfigure}[b]{0.33\linewidth}
\centering
\includegraphics[width=\linewidth]{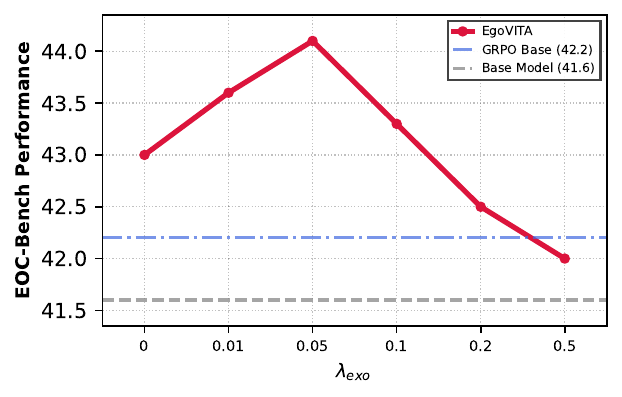}
\caption{EOC-Bench}

\end{subfigure}
\hspace{0.1cm}
\begin{subfigure}[b]{0.33\linewidth}
\centering
\includegraphics[width=\linewidth]{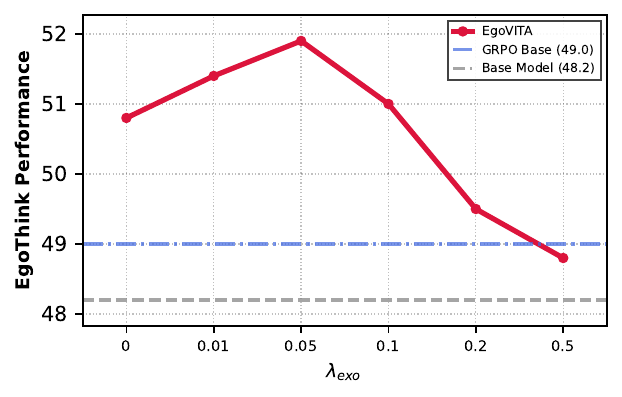}
\caption{EgoThink}

\end{subfigure}
\caption{\textbf{Effect of exocentric regularization strength $\lambda_{\mathrm{exo}}$ on egocentric benchmarks.} Moderate regularization ($\lambda_{\mathrm{exo}}{=}0.05$) yields peak egocentric performance across all four benchmarks, while stronger values degrade it.}
\label{fig:exo_regularization_ego_perfomance_ablation}
\end{figure*}

%% file: tables/vision_encoder.tex
\begin{table}[t!]
\centering
\caption{\textbf{Ablation on Trainable Vision Components.} We analyze training the vision encoder (VE) and vision projector (MLP) with DoRA adapters. The LLM is always trained during GRPO.}
\label{tab:component_ablation}
\begin{adjustbox}{width=\columnwidth,center}
\renewcommand{\arraystretch}{1.2}
\fontsize{6pt}{7pt}\selectfont
\setlength{\tabcolsep}{1mm}
\begin{tabular}{cccc cccc}
\toprule
 & \multicolumn{3}{c}{\textbf{Components Trained}} & \multicolumn{4}{c}{\textbf{Egocentric Benchmarks}} \\
\cmidrule(lr){2-4} \cmidrule(lr){5-8}
\textbf{Model} & \textbf{VE} & \textbf{MLP} & \textbf{LLM} & \textbf{EgoBlind} & \textbf{EgoPlan} & \textbf{EgoThink} & \textbf{EOC-Bench} \\
\midrule
Qwen2.5-VL-7B & \xmarkr & \xmarkr & \xmarkr & 29.7 & 30.2 & 48.2 & 41.6 \\
\midrule
Frozen Vision Pipeline & \xmarkr & \xmarkr & \cmarkg & 32.1 $\scriptstyle\textcolor{darkgreen}{\textbf{(+2.4)}}$ & 30.8 $\scriptstyle\textcolor{darkgreen}{\textbf{(+0.6)}}$ & 48.9 $\scriptstyle\textcolor{darkgreen}{\textbf{(+0.7)}}$ & 42.0 $\scriptstyle\textcolor{darkgreen}{\textbf{(+0.4)}}$ \\
Adaptive Projector & \xmarkr & \cmarkg & \cmarkg & 34.2 $\scriptstyle\textcolor{darkgreen}{\textbf{(+4.5)}}$ & 31.5 $\scriptstyle\textcolor{darkgreen}{\textbf{(+1.3)}}$ & 50.1 $\scriptstyle\textcolor{darkgreen}{\textbf{(+1.9)}}$ & 42.8 $\scriptstyle\textcolor{darkgreen}{\textbf{(+1.2)}}$ \\
Adaptive Encoder & \cmarkg & \xmarkr & \cmarkg & 35.8 $\scriptstyle\textcolor{darkgreen}{\textbf{(+6.1)}}$ & 32.1 $\scriptstyle\textcolor{darkgreen}{\textbf{(+1.9)}}$ & 50.8 $\scriptstyle\textcolor{darkgreen}{\textbf{(+2.6)}}$ & 43.2 $\scriptstyle\textcolor{darkgreen}{\textbf{(+1.6)}}$ \\
\rowcolor{PastaYellow}
\textbf{\modelname{} (Full)} & \cmarkg & \cmarkg & \cmarkg & 
\textbf{37.4} $\scriptstyle\textcolor{darkgreen}{\textbf{(+7.7)}}$ & 
\textbf{32.9} $\scriptstyle\textcolor{darkgreen}{\textbf{(+2.7)}}$ & 
\textbf{51.9} $\scriptstyle\textcolor{darkgreen}{\textbf{(+3.7)}}$ & 
\textbf{44.1} $\scriptstyle\textcolor{darkgreen}{\textbf{(+2.5)}}$ \\
\bottomrule
\end{tabular}
\end{adjustbox}
\end{table}

%% file: figures/acmg_temporal_grounding.tex
\begin{figure*}[!t]
    \centering
    \includegraphics[width=0.8\columnwidth]{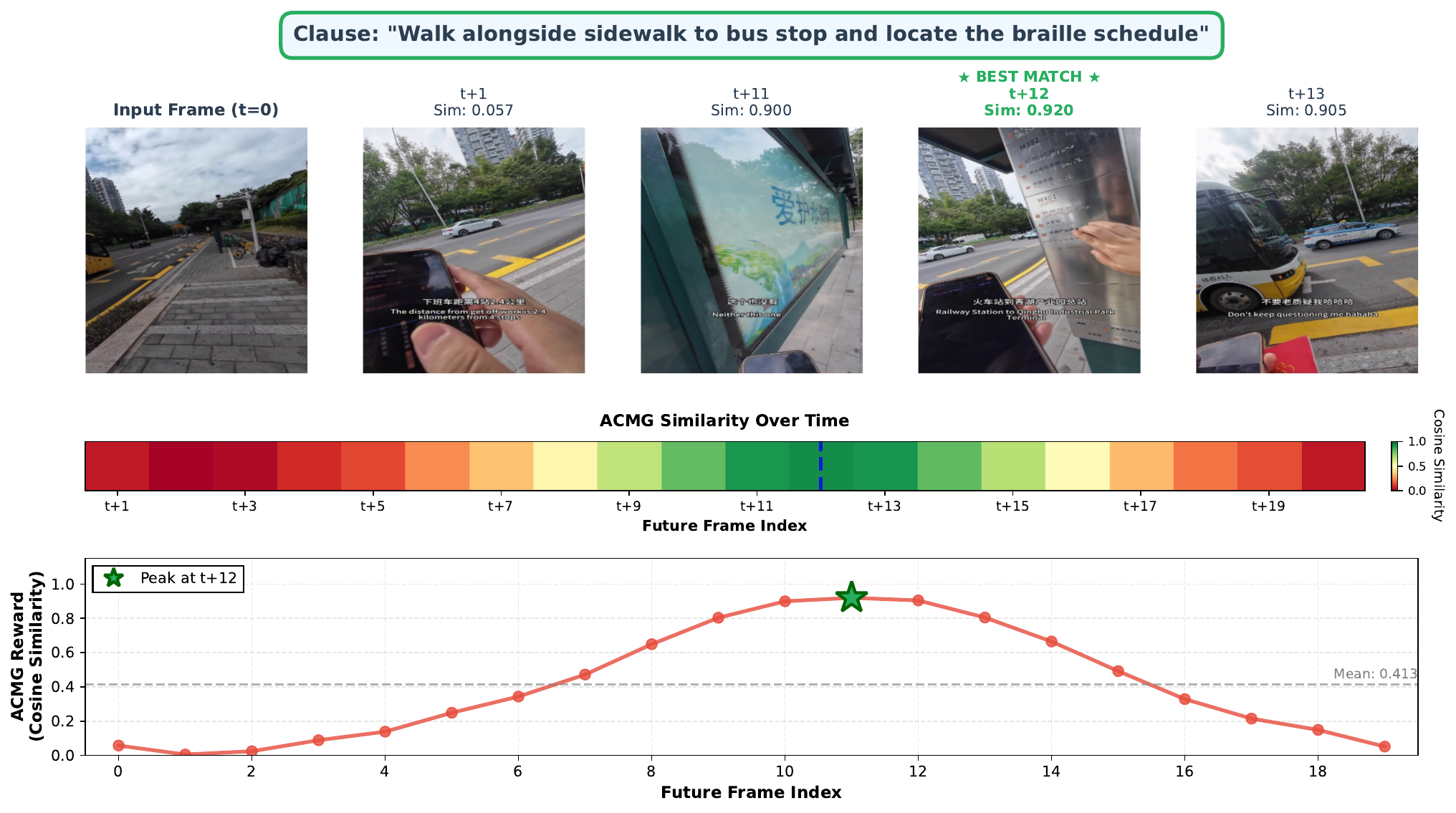}   
    \caption{\textbf{$R_{\mathrm{ACMG}}$ Temporal Grounding for a Single Clause.} Given the clause ``Walk alongside sidewalk to bus stop and locate the braille schedule,'' the Anticipation Head projects it into visual space and computes cosine similarity with future frames. The best match occurs at $t+12$ (Sim: 0.920) when the braille sign appears. The heatmap and curve show similarity peaks at the temporally correct frame, demonstrating that $R_{\mathrm{ACMG}}$ rewards clauses that accurately predict future visual states.}
    \label{fig:acmg_grounding}
\end{figure*}

%% file: tables/frames_ablation.tex
\begin{table}[t!]
\centering
\caption{\textbf{Ablation on Anticipation Window ($N$).} We analyze the sensitivity of the ACMG reward to $N$, the number of future frames used for grounding. Results on Qwen2.5-VL-7B. $N=16$ provides the best balance, as smaller windows are too myopic to capture delayed actions, while larger windows introduce noisy, spurious correlations.}
\label{tab:n_frames_ablation}
\begin{adjustbox}{width=0.7\columnwidth,center}
\renewcommand{\arraystretch}{1.2}
\fontsize{6pt}{7pt}\selectfont
\setlength{\tabcolsep}{1mm}
\begin{tabular}{l cccc}
\toprule
 & \multicolumn{4}{c}{\textbf{Egocentric Benchmarks}} \\
\cmidrule(lr){2-5}
\textbf{$N$ (Future Frames)} & \textbf{EgoBlind} & \textbf{EgoPlan} & \textbf{EgoThink} & \textbf{EOC-Bench} \\
\midrule
Qwen2.5-VL-7B & 29.7 & 30.2 & 48.2 & 41.6 \\
\midrule
$N=4$ (Too Myopic) & 32.0 $\scriptstyle\textcolor{darkgreen}{\textbf{(+2.3)}}$ & 31.0 $\scriptstyle\textcolor{darkgreen}{\textbf{(+0.8)}}$ & 49.3 $\scriptstyle\textcolor{darkgreen}{\textbf{(+1.1)}}$ & 42.4 $\scriptstyle\textcolor{darkgreen}{\textbf{(+0.8)}}$ \\
$N=8$ & 35.1 $\scriptstyle\textcolor{darkgreen}{\textbf{(+5.4)}}$ & 32.1 $\scriptstyle\textcolor{darkgreen}{\textbf{(+1.9)}}$ & 50.8 $\scriptstyle\textcolor{darkgreen}{\textbf{(+2.6)}}$ & 43.4 $\scriptstyle\textcolor{darkgreen}{\textbf{(+1.8)}}$ \\
\rowcolor{PastaYellow}
\textbf{$N=16$ (\modelname{})} & 
\textbf{37.4} $\scriptstyle\textcolor{darkgreen}{\textbf{(+7.7)}}$ & 
\textbf{32.9} $\scriptstyle\textcolor{darkgreen}{\textbf{(+2.7)}}$ & 
\textbf{51.9} $\scriptstyle\textcolor{darkgreen}{\textbf{(+3.7)}}$ & 
\textbf{44.1} $\scriptstyle\textcolor{darkgreen}{\textbf{(+2.5)}}$ \\
$N=32$ (Too Noisy) & 36.2 $\scriptstyle\textcolor{darkgreen}{\textbf{(+6.5)}}$ & 32.5 $\scriptstyle\textcolor{darkgreen}{\textbf{(+2.3)}}$ & 51.3 $\scriptstyle\textcolor{darkgreen}{\textbf{(+3.1)}}$ & 43.7 $\scriptstyle\textcolor{darkgreen}{\textbf{(+2.1)}}$ \\
\bottomrule
\end{tabular}
\end{adjustbox}
\end{table}

%% file: figures/acmg_box.tex
\begin{figure}[!t]
\centering
\includegraphics[width=0.7\columnwidth]{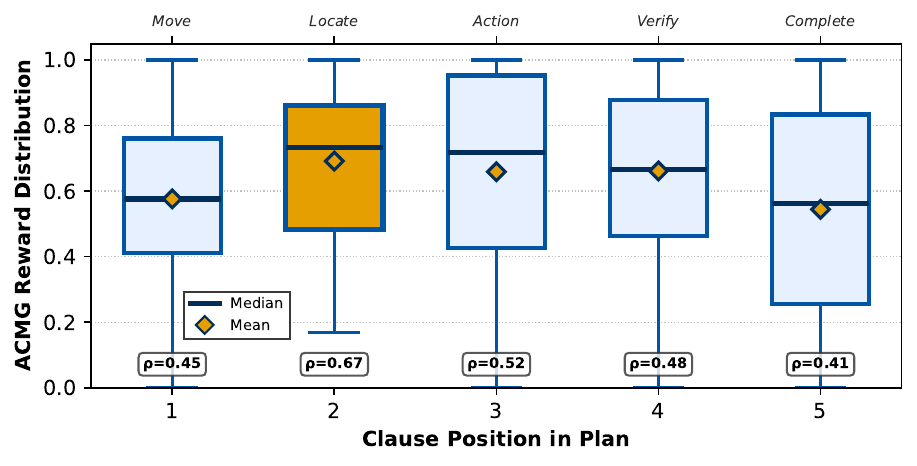}
\caption{\textbf{ACMG Reward Distribution by Clause Position.} 
Box plots show ACMG reward distributions for the first five clauses in \protect\egoplan sequences. 
Top labels indicate typical semantic roles at each position. 
Middle clauses (positions $2-3$) achieve higher rewards and stronger correlation ($\rho$) with answer correctness, 
demonstrating that ACMG  distinguishes between concrete action steps and abstract reasoning phases.}
\label{fig:clause_position}
\end{figure}

%% file: tables/qwen3vl_teacher.tex
\begin{table}[t!]
\centering
\caption{\textbf{EgoVITA with Qwen3-VL as teacher.}}
\label{tab:qwen3vl_based}
\begin{adjustbox}{width=\columnwidth,center}
\renewcommand{\arraystretch}{1.2}
\fontsize{6pt}{7pt}\selectfont
\setlength{\tabcolsep}{1mm}
\begin{tabular}{lcccc|ccc}
\toprule
& \multicolumn{4}{c}{\textbf{Egocentric Video Reasoning}} & \multicolumn{3}{c}{\textbf{Exocentric Video Understanding}} \\
\cmidrule(lr){2-5}\cmidrule(lr){6-8}
\textbf{Variant} & \textbf{Ego-Exo-Bench} & \textbf{EgoBlind} & \textbf{EgoPlan} & \textbf{EOC-Bench} & \textbf{MVBench} & \textbf{Video-MME} & \textbf{LVBench} \\
\midrule
Qwen2.5-VL & 31.7 & 29.7 & 30.2 & 41.6 & 65.2 & 62.2 & 42.8 \\
SFT & 33.2 & 34.8 & 31.3 & 41.8 & 63.2 & 60.1 & 40.2 \\
GRPO (Format + Answer) & 33.8 & 35.6 & 31.4 & 42.2 & 66.4 & 62.9 & 40.9 \\
\rowcolor{PastaYellow}
\textbf{EgoVITA} & 
\textbf{34.6} $\scriptstyle\textcolor{darkgreen}{\textbf{(+2.9)}}$ & 
\textbf{38.5} $\scriptstyle\textcolor{darkgreen}{\textbf{(+8.8)}}$ & 
\textbf{33.2} $\scriptstyle\textcolor{darkgreen}{\textbf{(+3.0)}}$ & 
\textbf{45.4} $\scriptstyle\textcolor{darkgreen}{\textbf{(+3.8)}}$ & 
\textbf{66.7} $\scriptstyle\textcolor{darkgreen}{\textbf{(+1.5)}}$ & 
\textbf{64.8} $\scriptstyle\textcolor{darkgreen}{\textbf{(+2.6)}}$ & 
\textbf{43.7} $\scriptstyle\textcolor{darkgreen}{\textbf{(+0.9)}}$ \\
\bottomrule
\end{tabular}
\end{adjustbox}
\end{table}

%% file: figures/qual_supp.tex
\begin{figure*}[!t]
\centering
\includegraphics[width=\textwidth]{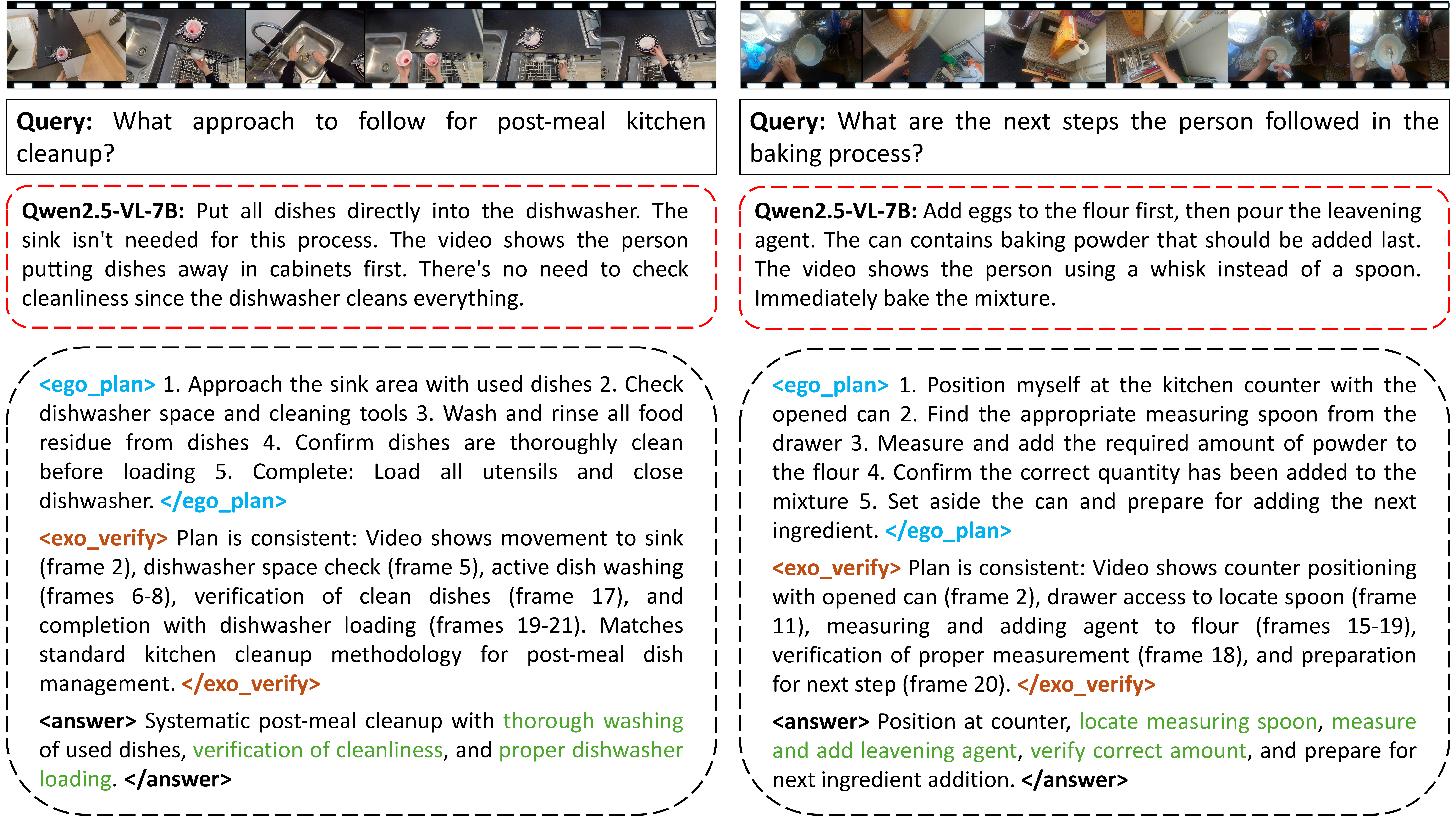}
\caption{\textbf{\modelname{} Qualitative Examples.}The baseline Qwen2.5-VL-7B~\cite{qwen2_5vl} shows factual errors and temporal hallucinations (e.g., inventing ``cabinet storage'' steps or objects like ``eggs'' and ``whisks''). In contrast, \modelname{} uses \protect\egoplan to generate coherent action sequences and \protect\exoverify to validate them against relevant video frames, producing visually grounded and temporally consistent responses.}

\label{fig:qualitative_supp}
\end{figure*}

%% file: figures/exo_fixing.tex
\begin{figure*}[!t]
\centering
\includegraphics[width=\textwidth]{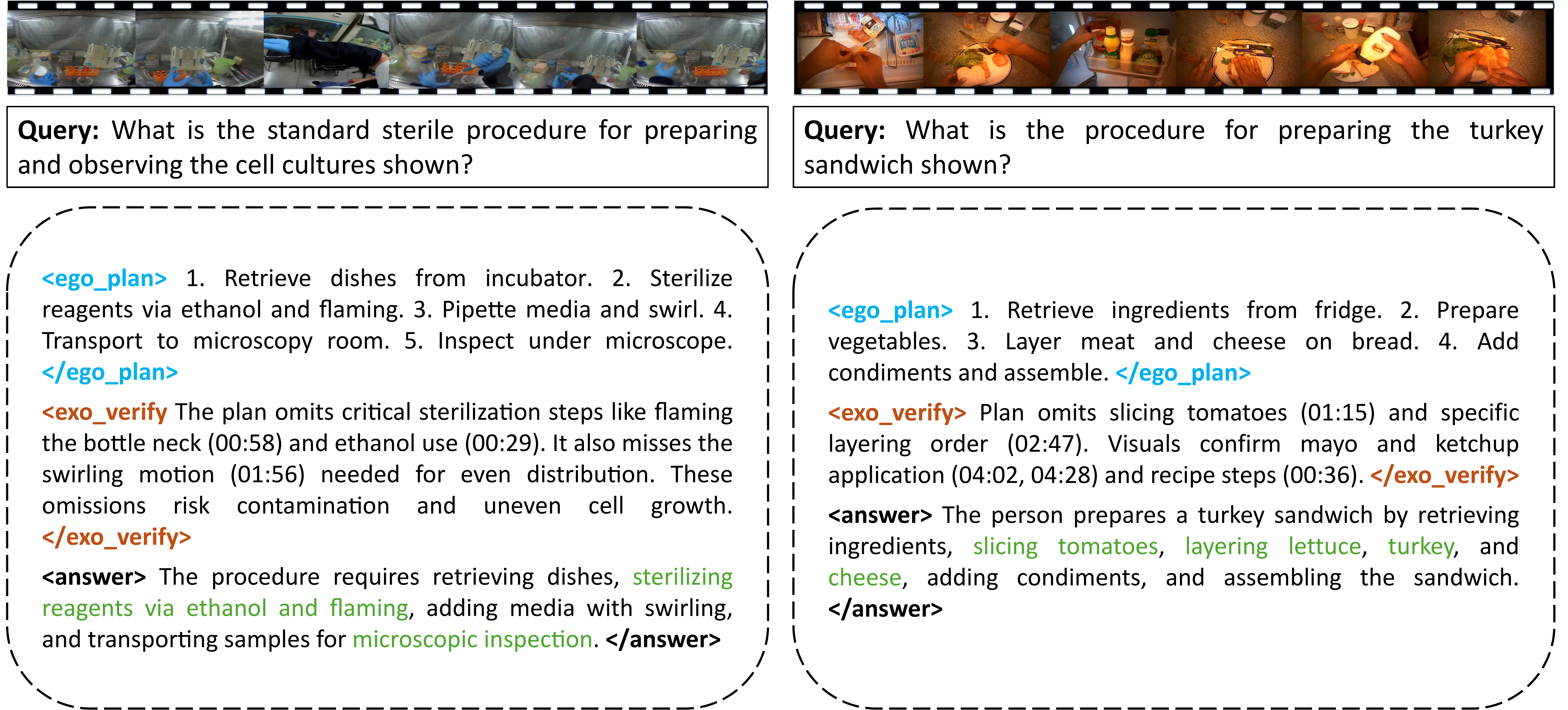}
\caption{\textbf{Exocentric verification identifies flaws in egocentric plans.} In both examples, \protect\exoverify detects omitted steps and incorrect orderings in the \protect\egoplan (e.g., missing sterilization steps in the lab procedure, missing slicing order in sandwich preparation). By reasoning from a third-person perspective, the model corrects these gaps, producing a more complete and grounded \protect\answer.}
\label{fig:exo_fixing}
\end{figure*}

%% file: figures/failure_analysis.tex
\begin{figure}[t!]
    \centering
    \includegraphics[width=0.65\columnwidth]{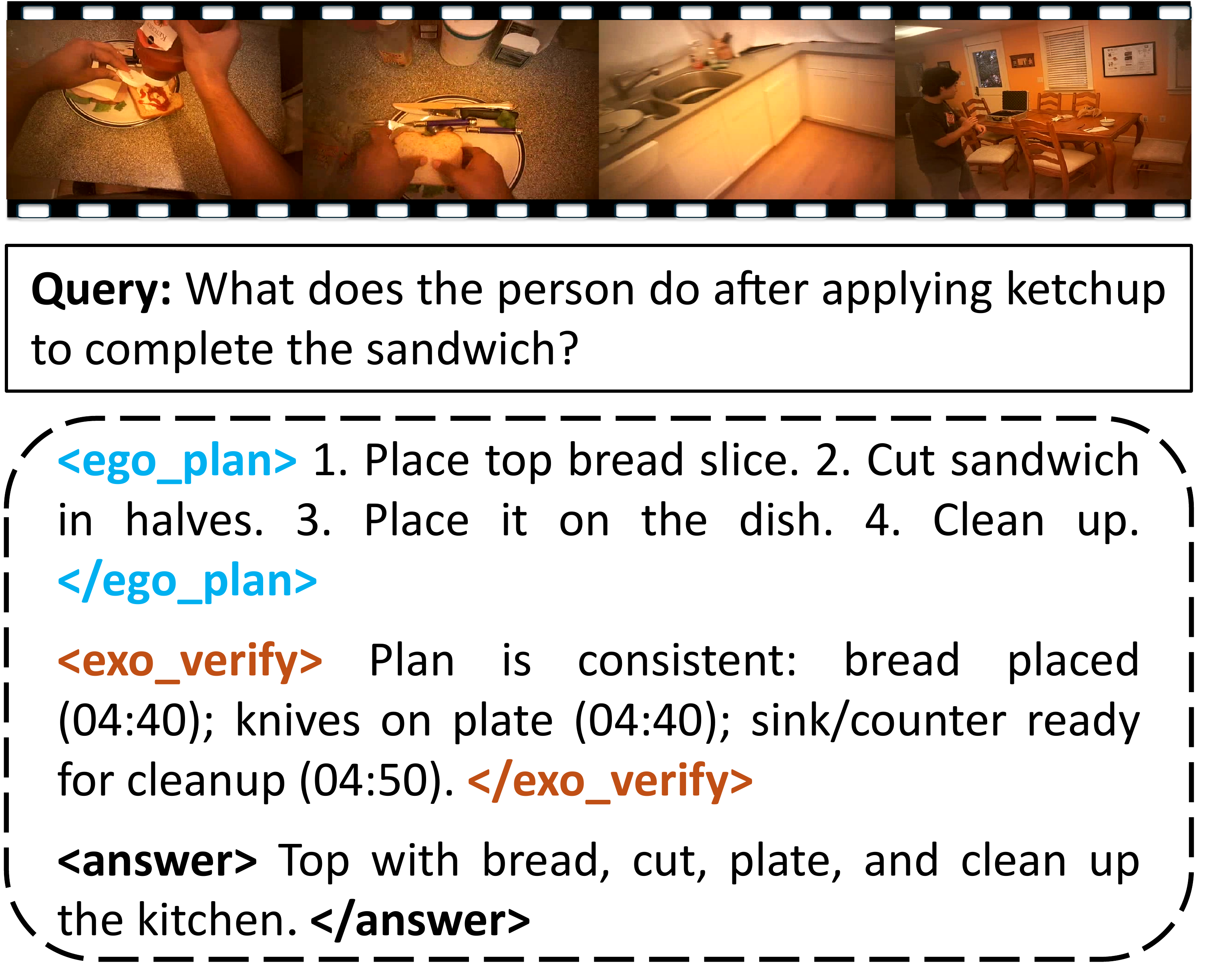}
    \caption{\textbf{Failure case.} After the person exits, the sink remains visible within the $N{=}16$ window. ACMG rewards the incorrect ``cleanup'' from persistent object co-occurrence. Maps to Types C+F in Table~\ref{tab:rollout_analysis}.}
    \label{fig:failure_case}
\end{figure}

\begin{table}[t!]
\centering
\caption{\textbf{Rollout failure analysis.} 3,278 rollouts tagged by Gemini-3.1-Flash.}
\label{tab:rollout_analysis}
\renewcommand{\arraystretch}{1.1}
\fontsize{7pt}{8pt}\selectfont
\setlength{\tabcolsep}{3mm}
\begin{tabular}{l r r}
\toprule
\textbf{Type} & \textbf{N} & \textbf{\%} \\
\midrule
A. All correct   & 1847 & 56 \\
B. Plan fixed by \textit{exo\_verify}    &  487 & 15 \\
C. Incorrect plan confirmed   &  312 & 10 \\
D. Fix attempted but failed    &  194 &  6 \\
\midrule
E. ACMG noisy ($N{=}16$ window)    &  341 & 10 \\
F. ACMG static feature bias     &   97 &  3 \\
\bottomrule
\end{tabular}
\end{table}

%% file: tables/arch.tex
\begin{table}[!t]
\centering
\caption{\textbf{Anticipation Head Architecture.} The MLP projects text embeddings ($D_{llm}$) to the visual space ($D_{vis}$).}
\label{tab:mlp_arch}
\begin{adjustbox}{width=0.5\textwidth,center}
\renewcommand{\arraystretch}{1.1}
\small
\begin{tabular}{l l l}
\toprule
\textbf{Stage} & \textbf{Operation} & \textbf{Output Shape} \\
\midrule
Input & -- & $B \times D_{llm}$ \\
Layer 1 & Linear Projection & $B \times D_{hidden}$ \\
Activation & GELU & $B \times D_{hidden}$ \\
Normalization & LayerNorm & $B \times D_{hidden}$ \\
Layer 2 & Linear Projection & $B \times D_{vis}$ \\
\bottomrule
\end{tabular}
\end{adjustbox}
\end{table}

%% file: figures/ego_plan.tex
\begin{figure*}[!t]
    \centering
    \begin{tcolorbox}[
        title={Prompt for Egocentric Planning SFT Data Generation (EgoProceL)},
        colframe=black,
        colback=gray!10,
        coltitle=white,
        fonttitle=\bfseries\small,
        width=\textwidth,
        boxsep=3pt,
        top=3pt,
        bottom=3pt,
        left=4pt,
        right=4pt
    ]
    \small
    You are an expert in egocentric video understanding and procedural planning. You will be provided with a list of timestamped atomic actions annotated from a first-person video. Your task is to convert these raw annotations into a high-quality instruction-following training sample containing a query, options, and a grounded plan.

    \textbf{Input Data:} A list of timestamped actions (Start Time, End Time, Action Description).

    \textbf{Your Task:}
    \vspace{-1mm}
    \begin{enumerate}[itemsep=1pt, parsep=0pt, topsep=2pt, leftmargin=12pt]
        \item \textbf{Synthesize a User Query:} Infer the overall goal of the video (e.g., ``How do I make brownies?'') and formulate it as a question a user would ask for assistance.
        \item \textbf{Generate Answer Options:} Create 4 distinct multiple-choice options (A, B, C, D) that summarize the procedure:
        \item \textbf{One Correct Option:} An accurate summary of the sequential steps.
        \item \textbf{Three Distractors:} Plausible but incorrect summaries (e.g., incorrect order, missing critical steps, or describing a different but related task).
        \item \textbf{Construct the Response:} Generate the target output using the following strictly enforced format:
        \item \texttt{<ego\_plan>}: A numbered, step-by-step plan. Each step must cite the specific timestamp interval from the input to ensure temporal grounding.
        \item \texttt{}: The final answer containing the correct option letter and its full text.
    \end{enumerate}
    \vspace{-1mm}
    \textbf{Input Annotations:} \{50.48-54.7: break egg\}, \{64.19-76.3: mix eggs\}, \{87.13-91.1: add water\}, ...

    \textbf{Target Output Format:}
    \vspace{-2mm}
    \begin{verbatim}
[User Query]: {Generated Query}
[Options]:  (A) {Option Text}  (B) {Option Text} ...
[Response]:
<ego_plan>
1. Break the egg into the bowl to start the batter (50.48 - 54.7).
2. Mix the eggs thoroughly using a whisk (64.19 - 76.3).
3. Add water to the mixture as specified (87.13 - 91.1). ...
</ego_plan>
<answer> (A) {Correct Option Text} </answer>
    \end{verbatim}
    \vspace{-3mm}
    \end{tcolorbox}
    \vspace{-2mm}
    \caption{Prompt used to transform raw temporal annotations into structured SFT samples for egocentric planning. This prompts the teacher model (Qwen2.5-VL-72B) to generate a grounded \protect\egoplan and a correct \protect\answer.}
    \label{fig:sft-prompt}
\end{figure*}

%% file: figures/exo_verify.tex
\begin{figure*}[!t]
    \centering
    \begin{tcolorbox}[
        title={Prompt for Exocentric Verification SFT Data Generation (HD-EPIC)},
        colframe=black,
        colback=gray!10,
        coltitle=white,
        fonttitle=\bfseries\small,
        width=\textwidth,
        boxsep=3pt,
        top=3pt,
        bottom=3pt,
        left=4pt,
        right=4pt
    ]
    \small
    You are an intelligent video reasoning assistant. You are provided with an egocentric video and a set of fine-grained, timestamped action annotations. Your goal is to synthesize a reasoning chain that not only plans the actions but also \textbf{verifies} them against visual evidence.

    \textbf{Input Data:}
    \vspace{-1mm}
    \begin{enumerate}[itemsep=1pt, parsep=0pt, topsep=2pt, leftmargin=12pt]
        \item \textbf{Video Context:} An egocentric video clip.
        \item \textbf{Annotations:} A list of atomic actions with start/end timestamps (e.g., ``13.73-14.11: flip the bottom part of the coffee machine'').
    \end{enumerate}
    \vspace{-1mm}
    \textbf{Your Task:} Generate a structured output containing two distinct blocks:
    \vspace{-1mm}
    \begin{enumerate}[itemsep=1pt, parsep=0pt, topsep=2pt, leftmargin=12pt]
        \item \texttt{<ego\_plan>}: Convert the annotations into a coherent, first-person procedural plan. Merge closely related atomic actions into logical steps where appropriate.
        \item \texttt{<exo\_verify>}: Act as an external auditor. For the plan you just generated, provide a \textbf{visual verification} rationale. Explicitly cite visual evidence (objects, hand-object interactions, and state changes) that confirms the plan is actually occurring in the video.
    \end{enumerate}
    \vspace{-1mm}
    \textbf{Input Annotations:} \{14.2-14.55: Put mug under nozzle\}, \{14.81-15.08: Pick up coffee capsule\}, \{16.08-17.26: Slide capsule into gap\}...

    \textbf{Target Output Format:}
    \vspace{-2mm}
    \begin{verbatim}
[User Query]: How do I prepare the coffee machine for brewing?
[Response]:
<ego_plan>
1. Place the mug under the coffee machine nozzle (14.2 - 14.55).
2. Pick up the coffee capsule from the left countertop (14.81 - 15.08).
3. Insert the capsule into the machine's slot (16.08 - 17.26). ...
</ego_plan>
<exo_verify>
The plan is visually consistent. At 14.5s, the right hand is clearly
visible placing the white mug under the spout. The interaction with
the capsule is confirmed by the hand trajectory moving from the
counter (14.9s) to the machine's top slot (16.5s), followed by the
lever being pulled down to lock it.
</exo_verify>
    \end{verbatim}
    \vspace{-3mm}
    \end{tcolorbox}
    \vspace{-2mm}
    \caption{Prompt used to generate exocentric verification data from HD-EPIC~\cite{perrett2025hd} annotations for third-person plan verification.}
    \label{fig:verify-prompt}
\end{figure*}